\documentclass[preprint,12pt]{elsarticle}
\usepackage[left]{lineno}           

\usepackage{amssymb}
\usepackage{amsmath}
\usepackage{graphicx}
\usepackage{booktabs}
\usepackage{multirow}

\usepackage{tikz}
\usetikzlibrary{arrows.meta, positioning, shadows.blur, fit}
\usepackage{xcolor}
\newcommand{\rev}[1]{#1}

\usepackage[ruled,vlined]{algorithm2e} 

\usepackage{array}     
\usepackage{makecell}  
\journal{AIC}

\begin{document}
\begin{frontmatter}

\title{A Human-in-the-Loop Control Paradigm for Construction Robotics:
From VR-Guided Teleoperation to Semi-Autonomous Dual-Arm Operation}

\author[label1]{CHEN Jinhua\fnref{cofirst}}
\author[label1]{XU Xiaoji\fnref{cofirst}}
\author[label2]{LIAO Zhongyuan\corref{cor1}}
\author[label2]{Huang Zhenmin}
\author[label2]{Zhang Ye}
\author[label1]{XIANG Changying\corref{cor2}}

\fntext[cofirst]{These authors contributed equally to this work.}
\cortext[cor1]{Corresponding author: zy.liao@connect.ust.hk}
\cortext[cor2]{Corresponding author: changyingx@ust.hk}

\address[label1]{Division of Integrative Systems and Design, Hong Kong University of Science and Technology, Hong Kong}
\address[label2]{Hong Kong University of Science and Technology, Hong Kong, China}

\begin{abstract}
\rev{Construction tasks in hazardous and unstructured environments require both human adaptability and robotic precision, yet existing approaches often treat teleoperation and autonomy as separate control modes. This paper presents a human-in-the-loop control paradigm that enables structured transitions among VR-guided teleoperation, semi-autonomous shared control, and rule-based autonomous execution for contact-rich dual-arm construction tasks. The proposed system combines immersive VR guidance, a manually defined library of reusable skill primitives, and a shared-autonomy framework with explicit human--robot role allocation and command-priority rules. This integration allows operators to retain supervisory and corrective authority while delegating repetitive and precision-critical actions to the robot. The paradigm is validated on a dual-arm platform through representative façade-installation tasks. Results show that autonomous execution provides the best performance under well-calibrated nominal conditions, whereas semi-autonomous control improves resilience when environmental disturbances exceed the predefined autonomous pipeline's correction capability. Semi-autonomous control also substantially reduces subjective operator workload compared with full teleoperation. These findings demonstrate that effective construction-robot control should adapt human involvement to task and environmental uncertainty, using autonomy for repeatable operations and shared control for disturbance recovery.}
\end{abstract}


\end{frontmatter}

\section{Introduction}

The construction industry is undergoing a profound transformation driven by automation and robotic technologies. Rising labor costs, stringent safety regulations, and accelerated urbanization have intensified the demand for solutions that enhance productivity while reducing occupational risks. In this context, robotic systems are increasingly viewed not only as tools for task execution, but as enablers of new \emph{human--robot collaboration paradigms} capable of operating safely and effectively in complex construction environments~\cite{liu2024robotics}.

Many construction operations, including structural assembly, surface maintenance, component installation, and on-site inspection, require dexterous dual-arm manipulation~\cite{atanasova2025building} comparable to human bimanual coordination. These tasks are typically performed in unstructured and hazardous settings such as high façades, outdoor scaffolds, and mechanically or thermally unstable sites. \rev{The present study is motivated by such high-altitude façade work, but all physical validation is conducted in a controlled laboratory mock-up rather than at height.} Prolonged manual operation under such conditions increases physical fatigue and accident risk, highlighting the need for robotic systems that can combine precise manipulation with adaptability and operational reliability~\cite{wei2023intelligent}.

While industrial robots have achieved a high level of autonomy in structured factory environments~\cite{rodriguez2021human}, directly transferring these capabilities to construction sites remains challenging. Fully autonomous control is often impractical due to uncertain geometries, variable material properties, and accumulated perception and localization errors. At the opposite extreme, pure teleoperation imposes a high cognitive burden on operators and scales poorly to multi-stage or bimanual tasks. These limitations motivate \emph{human-in-the-loop} approaches that explicitly integrate human spatial reasoning and decision-making with robotic precision and repeatability~\cite{gottardi2022shared}.

Despite recent progress in shared autonomy, two critical challenges remain. First, it is difficult to design autonomous control strategies that are sufficiently robust and generalizable to handle diverse contact-rich construction tasks such as drilling, screw fastening, and bracket assembly in dynamic environments. Second, effective construction robotics requires a control paradigm that supports \emph{transparent, safe, and flexible transitions} between human guidance and autonomous execution, rather than treating teleoperation and autonomy as disjoint modes.

To address these challenges, this paper proposes a \emph{human-in-the-loop control paradigm} for construction robotics that enables a structured transition from VR-guided teleoperation to semi-autonomous dual-arm operation. Within this paradigm, immersive VR interfaces provide intuitive human guidance and situational awareness during complex or uncertain task phases, while recurrent task patterns are encapsulated into autonomous skill primitives. A shared-autonomy framework allows smooth blending and switching between human input and autonomous execution, enabling reliable handling of critical contact phases in high-risk construction operations.

\rev{The main contributions of this work are summarized as follows:}
\begin{itemize}
    \item \rev{A task-level human/autonomy allocation framework for multi-stage, contact-rich dual-arm construction operations, enabling structured transitions between VR-guided teleoperation, semi-autonomous shared control, and rule-based autonomous execution under predefined conditions.}
    \item \rev{A dual-arm robotic system integrating immersive VR guidance with a manually defined rule-based skill library for drilling, bracket placement, and screw fastening.}
    \item \rev{A shared-autonomy formulation with skill-level control weights, explicit command-priority rules, and Subtask Allocation Ratio (SAR) analysis for quantifying human--robot role allocation.}
    \item \rev{A physical evaluation of drilling, bracket pick-and-place, and
    dual-arm screw fastening, including \(N{=}10\) novice operators in TeleOp and
    Semi-Auto, unmanned rule-based Auto repetitions, environmental-robustness
    testing for drilling, and a drilling-task Raw NASA-TLX assessment of subjective workload.}

\end{itemize}

\section{Related Work}

Research in construction robotics has progressed along multiple tightly coupled directions, driven by the need to operate safely and effectively in unstructured, hazardous, and contact-rich environments. From a control perspective, existing work can be broadly examined through three interrelated themes: 
(1) human--robot collaboration under environmental uncertainty, 
(2) dual-arm dexterous manipulation for complex construction operations, and 
(3) immersive teleoperation and shared autonomy as mechanisms for integrating human intelligence with robotic execution. 
This section reviews these themes with a focus on how current approaches position human involvement and autonomy within the control loop, and identifies the gaps motivating a transitional human-in-the-loop control paradigm.

\subsection{Human--Robot Collaboration in Unstructured Construction Environments}

Construction sites present challenges that fundamentally differ from structured industrial settings, including cluttered layouts, unstable surfaces, uncertain object geometries, and continuously evolving human activity. A comprehensive review by Xu et al.~\cite{Xu2025} highlighted that despite advances in automation, limitations in perception robustness and cognitive adaptability continue to hinder fully autonomous deployment in real construction scenarios.

To address these challenges, several works have explored formal models of human--robot collaboration. Wu et al.~\cite{wu2022human} analyzed collaborative productivity using agent-based modeling, while Alirezazadeh and Alexandre~\cite{alirezazadeh2022dynamic} proposed dynamic scheduling frameworks to coordinate human and robotic workflows. These studies emphasize that effective construction robotics requires explicit consideration of human roles rather than their elimination.

Safety and human factors are particularly critical in high-risk environments such as façades, scaffolds, and confined spaces. Predictable robot behaviors have been shown to improve operator trust~\cite{chang2024toward}, while proxemic models dynamically regulate safety zones around workers~\cite{kim2024understanding}. Perception and situational awareness further underpin safe collaboration: intent prediction from human motion~\cite{Zhang2024_Action} and cognitive-state-aware interaction~\cite{Liu2024_Cognitive} all contribute to more responsive robotic behavior. Digital twin systems~\cite{LIN2025106147} and mixed-reality deep reinforcement learning-based collision avoidance frameworks~\cite{Li2024_JMSY} provide additional layers of monitoring and safety enforcement.

While these approaches advance collaborative operation, most treat autonomy as a discrete capability rather than a controllable degree. This limits their ability to support fluid transitions between human guidance and robotic execution during multi-stage construction tasks.

\subsection{Dual-Arm Manipulation for Contact-Rich Construction Tasks}

Many construction operations inherently require coordinated bimanual manipulation, particularly when handling large, heavy, or deformable components. Dual-arm systems have demonstrated improved stability and task feasibility compared to single-arm solutions. Ge et al.~\cite{ge2025dual} presented a cooperative dual-arm framework for assembling bulky objects, while You et al.~\cite{You2024_DEXBOT} introduced a general-purpose manipulation architecture enabling versatile tool use.

Learning-based approaches have further expanded dual-arm capabilities. Kim et al.~\cite{Kim2024_DualAction} proposed goal-conditioned imitation learning for bimanual actions, and Yu et al.~\cite{Yu2024_Cloud} developed hierarchical pipelines that transfer VR-demonstrated skills to physical robots. Task-specific studies have addressed deformable object handling~\cite{Zhou2024} and coordinated construction workflows~\cite{mitterberger2022tie}, demonstrating the growing maturity of dual-arm manipulation techniques.

Complementary platforms, including aerial manipulators capable of drilling or fastening~\cite{dautzenberg2023perching, ding2021design} and wall-interaction robots~\cite{xingyu2025control}, further highlight the complexity of construction operations under environmental uncertainty. However, across these systems, autonomous dual-arm control alone remains insufficiently robust, and pure teleoperation imposes substantial cognitive and physical demands on operators. These limitations underscore the need for control frameworks that allow humans to guide high-level intent while delegating execution stability to the robot.

\rev{Recent construction robotics research has also emphasized demonstration-based skill transfer, immersive interfaces, and worker-centered shared control. Yu et al.~\cite{Yu2024_Cloud} proposed a cloud-based hierarchical imitation-learning pipeline for transferring construction skills from human workers to assisting robots. Our contribution focuses on skill-level autonomy allocation and physical dual-arm validation for contact-rich façade tasks (drilling, bracket placement, and screw fastening).}

\subsection{Immersive Teleoperation and Shared Autonomy}

Immersive teleoperation interfaces, particularly those based on VR and digital twins, provide intuitive channels for human guidance while maintaining operator safety. Digital twin environments have been used for trajectory planning and monitoring in excavation~\cite{fu2023digital}, while in-situ virtual representations enhance spatial understanding in complex sites~\cite{Yoon2024_InSitu}. Collaborative VR systems further enable distributed human coordination~\cite{Liu2023}.

Interface design strongly influences operator performance. Optimized visual layouts reduce cognitive load~\cite{Chae2024}, and adaptive viewpoint control assists operators in managing complex scenes~\cite{Yoon2025}. Communication latency remains a fundamental challenge~\cite{Seo2024, Du2024_Delay}, motivating perceptual and control-level compensation strategies.

To mitigate the limitations of pure teleoperation, shared autonomy approaches blend human input with autonomous correction. Hybrid frameworks refine operator commands to stabilize execution~\cite{Phung2024_Shared}, demonstrating clear benefits in manipulation tasks. Nevertheless, most existing systems still conceptualize teleoperation and autonomy as separate or sequential modes. Such rigid separation limits their effectiveness in construction scenarios that require continuous adjustment of autonomy levels, especially during contact-rich and multi-stage operations.

\subsection{Summary and Research Gap}

In summary, prior work demonstrates substantial progress in human--robot collaboration, dual-arm manipulation, and immersive teleoperation for construction robotics. However, fully autonomous systems remain brittle in unstructured environments, while pure teleoperation imposes excessive cognitive demands on operators. Existing shared-autonomy approaches often lack mechanisms for smooth, context-aware transitions between human guidance and autonomous execution.

These limitations motivate a \emph{human-in-the-loop control paradigm} that treats teleoperation and autonomy as points along a continuous control spectrum. By enabling structured transitions from VR-guided teleoperation to semi-autonomous dual-arm operation, such a paradigm provides a principled foundation for safe, efficient, and scalable execution of complex construction tasks, bridging current system-level solutions and future learning-based autonomy.

\rev{Table~\ref{tab:positioning} positions the present work relative to representative prior construction robotics systems. Unlike prior studies that often focus on a single interface or single-arm validation, this work combines VR teleoperation, dual-arm physical validation, task-level shared autonomy, SAR-based role-allocation analysis, and novice-operator evaluation in contact-rich façade tasks.}

{\color{blue}
\begin{table}[t]
\centering
\caption{Positioning of the present work relative to representative robotic
systems relevant to construction automation, skill learning, dual-arm
assembly, and shared autonomy. Abbreviations: DT = digital twin; VR = virtual
reality; SA = shared autonomy; HIL = hierarchical imitation learning.
``Subjects'' denotes a formal human-subject evaluation, and ``Workload''
denotes whether subjective workload was directly measured.}
\label{tab:positioning}
\footnotesize
\renewcommand{\arraystretch}{1.15}
\setlength{\tabcolsep}{3pt}
\resizebox{\textwidth}{!}{%
\begin{tabular}{lllllllll}
\toprule
\textbf{Work} & \textbf{Interface} & \textbf{Robot} & \textbf{Arm(s)} & \textbf{Control} & \textbf{Skill method} & \textbf{Task} & \textbf{Subjects} & \textbf{Workload} \\
\midrule
Fu et al.~\cite{fu2023digital} & DT & Excavator & N/A & Auto & Trajectory generation & Mining excavation & No & No \\
Yu et al.~\cite{Yu2024_Cloud} & VR demo & KUKA KR120 & Single & LfD/Auto & HIL (GRU/VAE) & Ceiling installation & No & No \\
Ge et al.~\cite{ge2025dual} & N/A & Industrial & Dual & Auto & Cooperative force control & Multi-peg assembly & No & No \\
Phung et al.~\cite{Phung2024_Shared} & VR/language/gesture & Underwater ROV & Single & SA & Supervisory control & Underwater sampling & Yes & No \\
\textbf{This work} & \textbf{VR} & \textbf{FAIRINO} & \textbf{Dual} &
\textbf{TeleOp/SA/Auto} & \textbf{Rule-based library} &
\textbf{Drill/bracket/screw} &
\textbf{Yes} &
\textbf{Yes} \\
\bottomrule
\end{tabular}%
}
\end{table}
}


\section{System Overview}

\begin{figure}[htbp]
    \centering
    \includegraphics[width=1\linewidth]{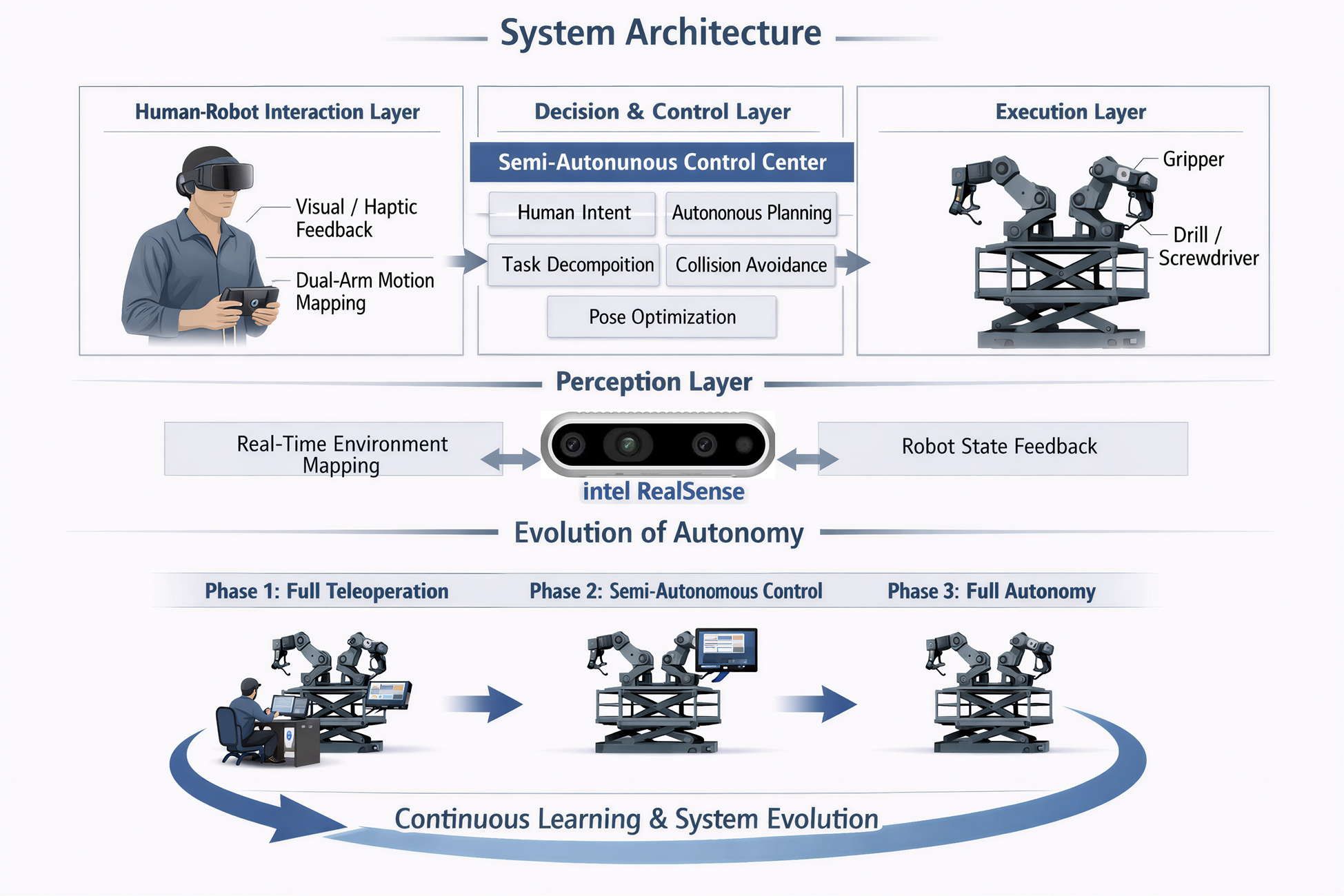}
    \caption{Overview of the proposed robotic system.}
    \label{fig:system}
\end{figure}

The proposed framework integrates immersive human guidance, modular
skill-level autonomy, and low-level robotic control within a closed-loop
architecture, as depicted in Fig.~\ref{fig:system}. Recurrent construction
subtasks are represented by manually defined skill primitives in a fixed
rule-based library. During execution, the operator can directly teleoperate
the robot, invoke autonomous skills, or provide corrective guidance in
shared-control states. The framework therefore supports structured
transitions among TeleOp, Semi-Auto, and rule-based Auto according to the
selected task and control allocation.

\subsection{Hardware Setup}

\begin{figure}[htbp]
    \centering
    \includegraphics[width=0.9\linewidth]{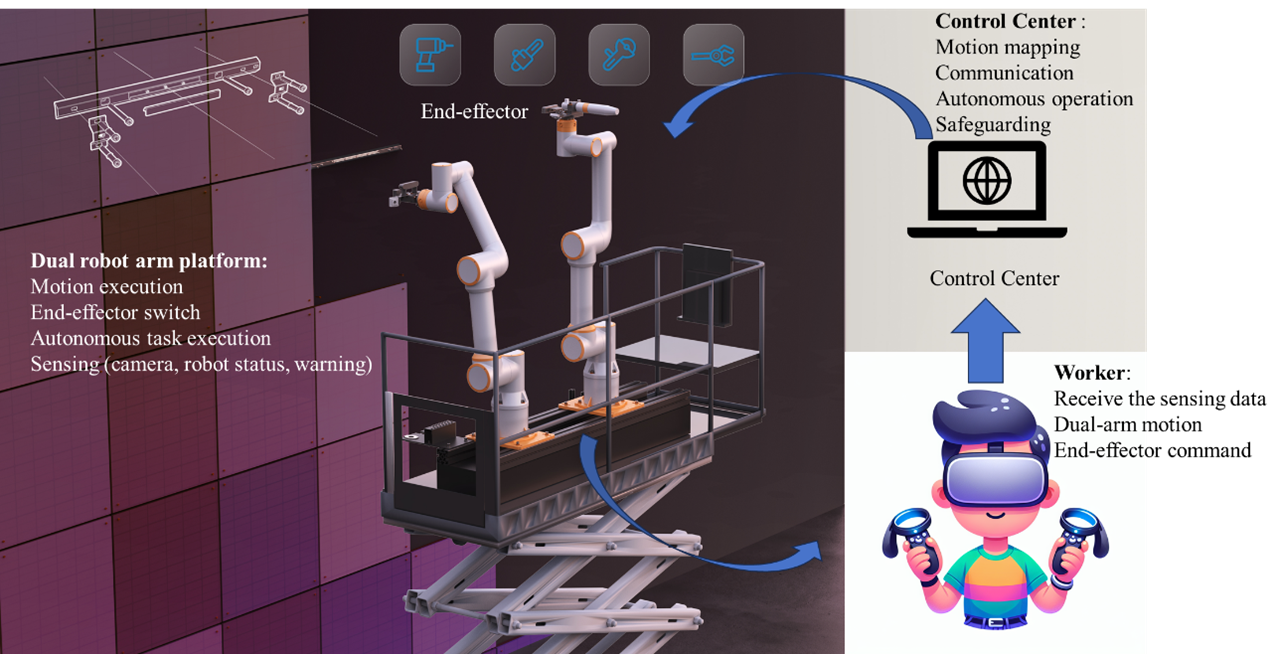}
    \caption{Hardware setup of the semi-autonomous dual-arm robotic platform.}
    \label{fig:Hardwarese}
\end{figure}

The experimental platform,  as shown in Fig.~\ref{fig:Hardwarese}, consists of three tightly integrated subsystems: a VR-based teleoperation interface, a dual-arm robotic system, and a centralized workstation. Together, they support immersive operator interaction, bimanual manipulation, autonomous skill execution, and synchronized system-level coordination. 

\begin{itemize}
\item \textbf{VR Teleoperation Interface.}
The operator interacts with the system using a Meta VR headset equipped with 6-DoF controller tracking, enabling accurate capture of hand poses and motion trajectories. Real-time RGB-D visual feedback provides situational awareness of the remote environment. Through this interface, the operator conveys high-level task intent, performs direct teleoperation when required, and triggers end-effector actions for fine-grained task adjustments.

\item \textbf{Dual-Arm Robotic System.}  
The robotic platform comprises two 6-DoF FAIRINO arms fitted with a set of interchangeable end-effectors, including a drilling module, an electric screwdriver, a suction gripper, and a general-purpose parallel jaw gripper. This configuration accommodates a diverse range of construction-oriented tasks. The robot executes coordinated bimanual motions, invokes autonomous skill primitives, performs tool switching, and continuously reports joint states, force signals, and system diagnostics to support shared autonomy.

\item \textbf{Central Workstation.}  
A unified computation and communication workstation handles motion retargeting between the operator and the robot workspace, high-frequency control communication, autonomous behavior planning, and global safety supervision. It integrates multimodal perception data, manages transitions between teleoperation and autonomous modes, and ensures synchronized, closed-loop operation across all hardware components.

\end{itemize}

\subsection{Human-in-the-Loop Semi-Autonomous Control Paradigm}

Construction operations such as maintenance, assembly, disassembly, drilling, fastening, and bracket installation are characterized by geometric uncertainty, structural variability, and contact-rich interactions. These properties make fully autonomous control brittle across diverse task instances, while pure teleoperation imposes excessive cognitive and operational burden on human operators. 

To address this gap, we propose a \textbf{human-in-the-loop semi-autonomous control paradigm} that explicitly treats autonomy as a \emph{continuous and adjustable spectrum} rather than a binary choice. Within this paradigm, human semantic reasoning and contextual judgment are preserved at the decision-making level, while robots assume responsibility for precise, repeatable, and safety-critical execution.

The proposed paradigm is designed to address three fundamental challenges:
\begin{itemize}
    \item translating high-level human intent into structured, robot-executable actions,
    \item allocating control authority between humans and robots in a task-aware and adaptive manner, and
    \item ensuring safe, efficient, and robust execution across varying levels of autonomy.
\end{itemize}

To operationalize this paradigm, the system is structured along two complementary dimensions:
\textbf{(I)} a multi-layered control structure that defines \emph{how responsibilities are distributed}, and
\textbf{(II)} a three-step task decomposition process that specifies \emph{how autonomy is enacted during execution}.

\subsubsection{Multi-layered Control Structure (Structural Dimension)}
\begin{figure}
    \centering
    \includegraphics[width=0.75\linewidth]{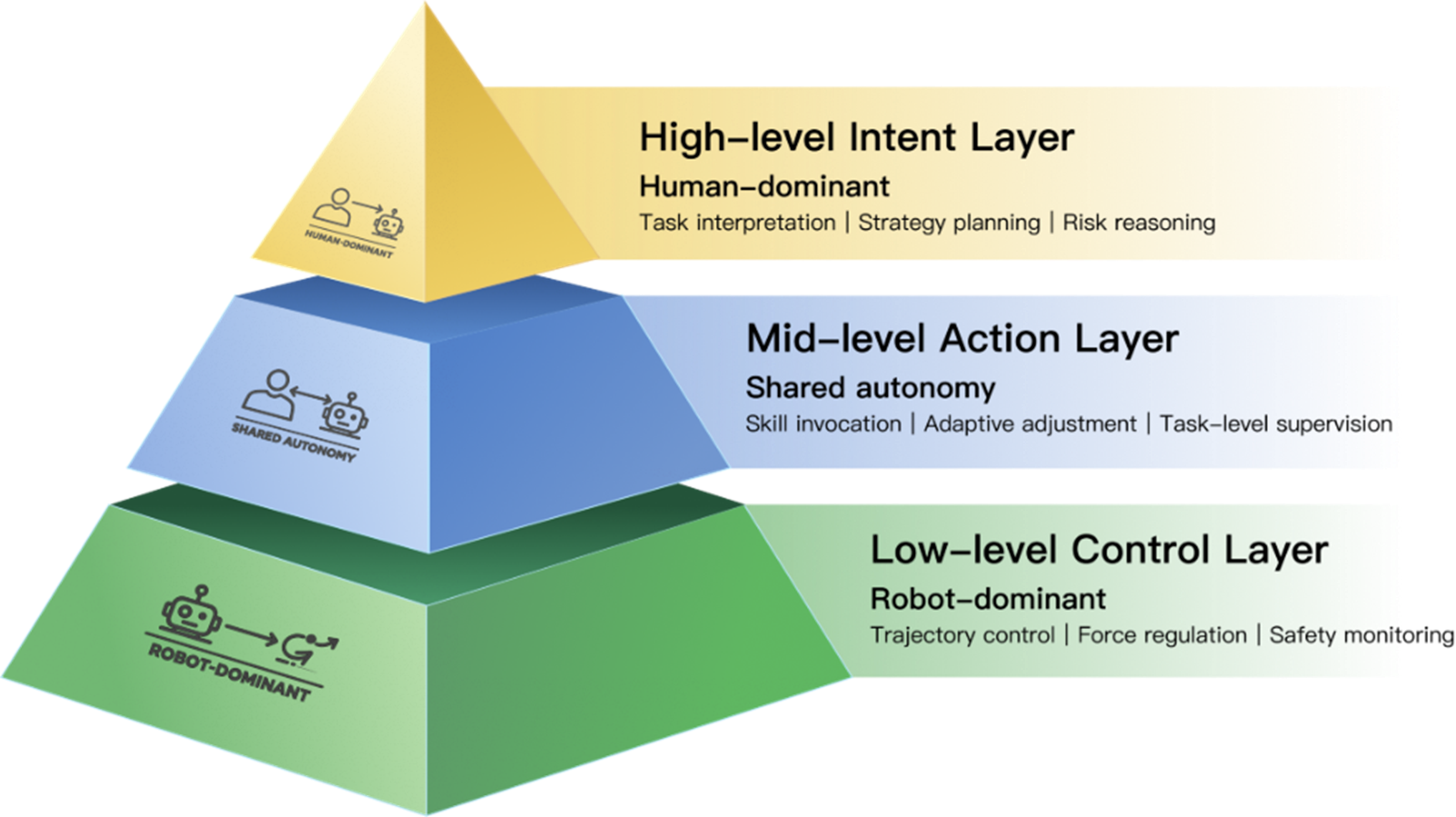}
    \caption{Three-layer structural architecture from human-dominant intent to robot-dominant control.}
    \label{fig:structure}
\end{figure}







The proposed architecture comprises three layers that explicitly organize control authority along a human--robot spectrum, as illustrated in Fig.~\ref{fig:structure}. Rather than enforcing rigid mode switching, the layers define a gradual transition from human-dominant decision-making to robot-dominant execution.

At the top of the hierarchy, the \textbf{High-level Intent Layer} is human-dominant and captures semantic reasoning, contextual interpretation, and strategic planning. In this layer, operators specify task goals, decompose operations (e.g., align--insert--fasten), assess environmental risks, and detect anomalous conditions. This layer preserves the irreplaceable strengths of human cognition in unstructured construction environments.

The \textbf{Mid-level Action Layer} serves as the core shared-autonomy interface. Human intent is transformed into robot-executable skill primitives, such as alignment, grasping, insertion, pressing, and tightening. While the human provides coarse guidance and strategic cues, the robot performs adaptive refinement using visual and force feedback. This layer enables continuous blending of human and robot contributions and constitutes the operational center of the human-in-the-loop paradigm.

The \textbf{Low-level Control Layer} is robot-dominant and executes high-frequency control loops, including trajectory tracking, hybrid force--position regulation, contact-state estimation, and safety monitoring. By autonomously handling these precision-critical components, the robot ensures stable and repeatable execution even under dynamic and uncertain conditions.

Together, these layers form a coherent control hierarchy in which autonomy is neither fixed nor binary, but dynamically allocated according to task context, uncertainty, and operator preference.

\subsubsection{Three-step Task Decomposition (Process Dimension)}

The execution of construction tasks under the proposed paradigm follows a three-step decomposition process that operationalizes the human-in-the-loop control spectrum at runtime.

Let the high-level task be represented by an intent vector $\mathbf{T}$, and let  
$\mathcal{S} = \{s_1, s_2, \dots, s_N\}$ denote the library of skill primitives.

The semi-autonomous workflow is decomposed into three steps.

\paragraph{Step 1: Task Intent Parsing}
The system derives a task sketch $\mathcal{K}$:
\begin{equation}
\mathcal{K} = f_\text{parse}(\mathbf{T}, \mathcal{C}),
\end{equation}
where $\mathcal{C}$ denotes scene context.  
The output includes:
\begin{itemize}
    \item candidate skill set $\mathcal{S}_c \subseteq \mathcal{S}$,
    \item responsibility map $R: \mathcal{S}_c \rightarrow \{\text{Human}, \text{Robot}\}$.
\end{itemize}

\paragraph{Step 2: Skill Selection and Role Allocation}
The operator selects an autonomy mode for each skill primitive:
\begin{equation}
M(s_i) \in \{\text{TeleOp},\, \text{Shared},\, \text{Auto}\}.
\end{equation}

\rev{
At each control cycle, the task state machine activates one skill
$s_{i_t}\in\mathcal{S}_c$, where $i_t$ denotes the index of the active skill
at time $t$. The autonomous controller generates the command
\begin{equation}
u_r^{i_t}(t)
=
f_{\mathrm{ctrl}}\!\left(x_t,s_{i_t},\mathcal{E}_t\right),
\end{equation}
where $x_t$ is the current robot state and $\mathcal{E}_t$ contains the
current perception, contact, and safety information.

For the active skill, the human and autonomous commands are blended as
\begin{equation}
u_{\mathrm{blend}}(t)
=
\alpha_{i_t}(t)u_h^{i_t}(t)
+
\beta_{i_t}(t)u_r^{i_t}(t),
\qquad
\alpha_{i_t}(t)+\beta_{i_t}(t)=1,
\label{eq:active_blending}
\end{equation}
where $u_h^{i_t}(t)$ and $u_r^{i_t}(t)$ denote the human and autonomous
commands for the active skill, respectively.
In the present implementation, $\alpha_i(t)$ and $\beta_i(t)$ are predefined at the skill and state-machine levels rather than learned or continuously adapted online. TeleOp assigns control primarily to the operator, rule-based Auto assigns control to the autonomous skill controller, and Shared uses predefined human--robot weight combinations for different execution states. The weights remain constant within each state and are updated only when a predefined state-transition condition is satisfied, such as reaching an alignment-error threshold, detecting stable contact, completing a skill stage, or receiving an operator-override command. Sensor confidence and contact-state checks are used to determine whether execution proceeds to the next state, returns control to the operator, or triggers a safe stop. Conflicting commands are resolved according to the priority rules in Table~\ref{tab:priority}.}
Early stages of a task typically favor higher human contribution for situational
understanding, target selection, and coarse guidance, while stable and repetitive
execution phases shift control toward the robot, progressively reducing operator workload.

\begin{table}[t]
\centering
\caption{\rev{Command-priority and conflict-resolution rules for shared-control execution.}}
\label{tab:priority}
\footnotesize
\resizebox{\textwidth}{!}{%
\begin{tabular}{clp{7.5cm}}
\toprule
\textbf{Priority} & \textbf{Source} & \textbf{Behavior} \\
\midrule
1 & Emergency stop / safety interlock & Immediate halt; all motion commands suppressed \\
2 & Workspace/force limits & Autonomous contact limits override operator motion in protected states \\
3 & Operator override (grip engaged) & Human command supersedes autonomous trajectory in shared states \\
4 & Autonomous skill controller & Default execution when no override is active \\
\bottomrule
\end{tabular}%
}
\end{table}

\paragraph{Continuous Subtask Allocation Ratio (SAR)}
To provide a quantitative measure of human involvement, we define the
\emph{Subtask Allocation Ratio (SAR)} as the average human control contribution across all
skill primitives in a task. Unlike time-based intervention measures, this continuous SAR
captures the effective human participation at the control level, independent of task
execution speed or transient fluctuations in control authority.

Formally, for a manipulation task decomposed into $N$ skill primitives
$\{\mathcal{S}_1, \mathcal{S}_2, \dots, \mathcal{S}_N\}$, with execution intervals
$[t_0^i, t_f^i]$ for each skill $s_i$, SAR is computed as:
\begin{equation}
\mathrm{SAR} = \frac{1}{N} \sum_{i=1}^{N} \overline{\alpha_i}, \quad
\overline{\alpha_i} = \frac{1}{t_f^i - t_0^i} \int_{t_0^i}^{t_f^i} \alpha_i(t) \, dt,
\end{equation}
where $\overline{\alpha_i}$ denotes the average human control weight for skill $s_i$. 
By construction, $\mathrm{SAR} \in [0,1]$, where values closer to 1 indicate greater human
involvement, and values closer to 0 indicate higher autonomy.

\rev{
\paragraph{Step 3: Shared-autonomy Execution}
The blended command in Eq.~\eqref{eq:active_blending} is passed through a
priority-based command-selection and safety layer. The final actuator command
is defined as
\begin{equation}
u_{\mathrm{final}}(t)
=
\begin{cases}
0,
& \text{if an E-stop or safety interlock is active},\\[1mm]
\Pi_{\mathrm{safe}}\!\left(u_h^{i_t}(t),\mathcal{E}_t\right),
& \text{if operator override is active},\\[1mm]
\Pi_{\mathrm{safe}}\!\left(u_{\mathrm{blend}}(t),\mathcal{E}_t\right),
& \text{otherwise},
\end{cases}
\label{eq:priority_command}
\end{equation}
where $\Pi_{\mathrm{safe}}(\cdot)$ denotes the safety filter that enforces
workspace, Cartesian-speed, force/torque, and contact-state limits. Depending
on the detected violation, the safety filter clips or rejects the candidate
command, stops the robot, or triggers a predefined retraction behavior.

Thus, operator override replaces the blended or autonomous command rather
than being added to it. The override remains subject to the same workspace,
speed, force, and emergency-stop constraints as autonomous execution.
}

The procedure is summarized in Algorithm~\ref{algo}.

\rev{
\begin{algorithm}[H]
\SetAlgoLined
\KwIn{Selected task $\mathbf{T}$, scene context $\mathcal{C}$}
\KwOut{Executed robot trajectory and skill actions}

\textbf{Step 1: Task and Skill Initialization}\;
$\mathcal{K},\mathcal{S}_c,R
\gets f_{\mathrm{parse}}(\mathbf{T},\mathcal{C})$\;

\textbf{Step 2: Skill and Autonomy Allocation}\;
\For{$s_i\in\mathcal{S}_c$}{
    $M(s_i)\gets$ select TeleOp / Shared / Auto\;
    Assign state-dependent weights $(\alpha_i,\beta_i)$\;
}

\textbf{Step 3: Priority-Based Execution}\;
\While{task not complete}{
    Sense robot state $x_t$ and environment $\mathcal{E}_t$\;
    Determine active skill $s_{i_t}$ from the task state machine\;

    $u_r^{i_t}(t)
    \gets f_{\mathrm{ctrl}}(x_t,s_{i_t},\mathcal{E}_t)$\;

    $u_{\mathrm{blend}}(t)
    \gets
    \alpha_{i_t}(t)u_h^{i_t}(t)
    +
    \beta_{i_t}(t)u_r^{i_t}(t)$\;

    \eIf{E-stop or safety interlock active}{
        $u_{\mathrm{final}}(t)\gets 0$\;
    }{
        \eIf{operator override active}{
            $u_{\mathrm{candidate}}(t)\gets u_h^{i_t}(t)$\;
        }{
            $u_{\mathrm{candidate}}(t)\gets u_{\mathrm{blend}}(t)$\;
        }
        $u_{\mathrm{final}}(t)
        \gets
        \Pi_{\mathrm{safe}}
        \!\left(u_{\mathrm{candidate}}(t),\mathcal{E}_t\right)$\;
    }

    Apply $u_{\mathrm{final}}(t)$ to the active robot controller\;
    Update the task state from robot and sensor feedback\;
}
\caption{Task decomposition, skill-level autonomy allocation, and
priority-based shared-control execution.}
\label{algo}
\end{algorithm}
}

\rev{The functions in Algorithm~\ref{algo} are implemented as follows.
The function $f_{\mathrm{parse}}(\mathbf{T},\mathcal{C})$ deterministically
maps the selected construction task and scene context to an ordered skill
sequence, arm/tool assignment, and state-machine configuration. The function
$f_{\mathrm{ctrl}}(x_t,s_{i_t},\mathcal{E}_t)$ generates the autonomous
motion or contact-control command for the currently active skill.

Operator override is implemented through a supervisory ROS flag activated
by the VR grip button. When the flag is active, the human command replaces
the blended or autonomous command before safety filtering. The emergency
stop is implemented as a separate hardware and software safety interlock; it
does not inject an override command but instead suppresses all actuator
commands. Workspace, speed, and force/torque limits are applied to both human
and autonomous candidate commands through
$\Pi_{\mathrm{safe}}(\cdot)$.}


In summary, the proposed human-in-the-loop semi-autonomous control paradigm integrates immersive VR-based intention input, modular skill-level autonomy, and precise low-level robotic control into a unified framework. The multi-layered structure clarifies the distribution of responsibilities across the control spectrum, while the three-step task decomposition provides a principled mechanism for translating high-level intent into safe and efficient dual-arm execution.

By treating teleoperation, shared control, and autonomous execution as points along a continuous control spectrum, this paradigm enables robust performance in complex construction environments and establishes a learning-ready foundation for future transitions toward higher levels of autonomy.

\section{Case Study: Dual-Arm Semi-Autonomous \rev{Bracket Installation}}

\begin{figure}
    \centering
    \includegraphics[width=1\linewidth]{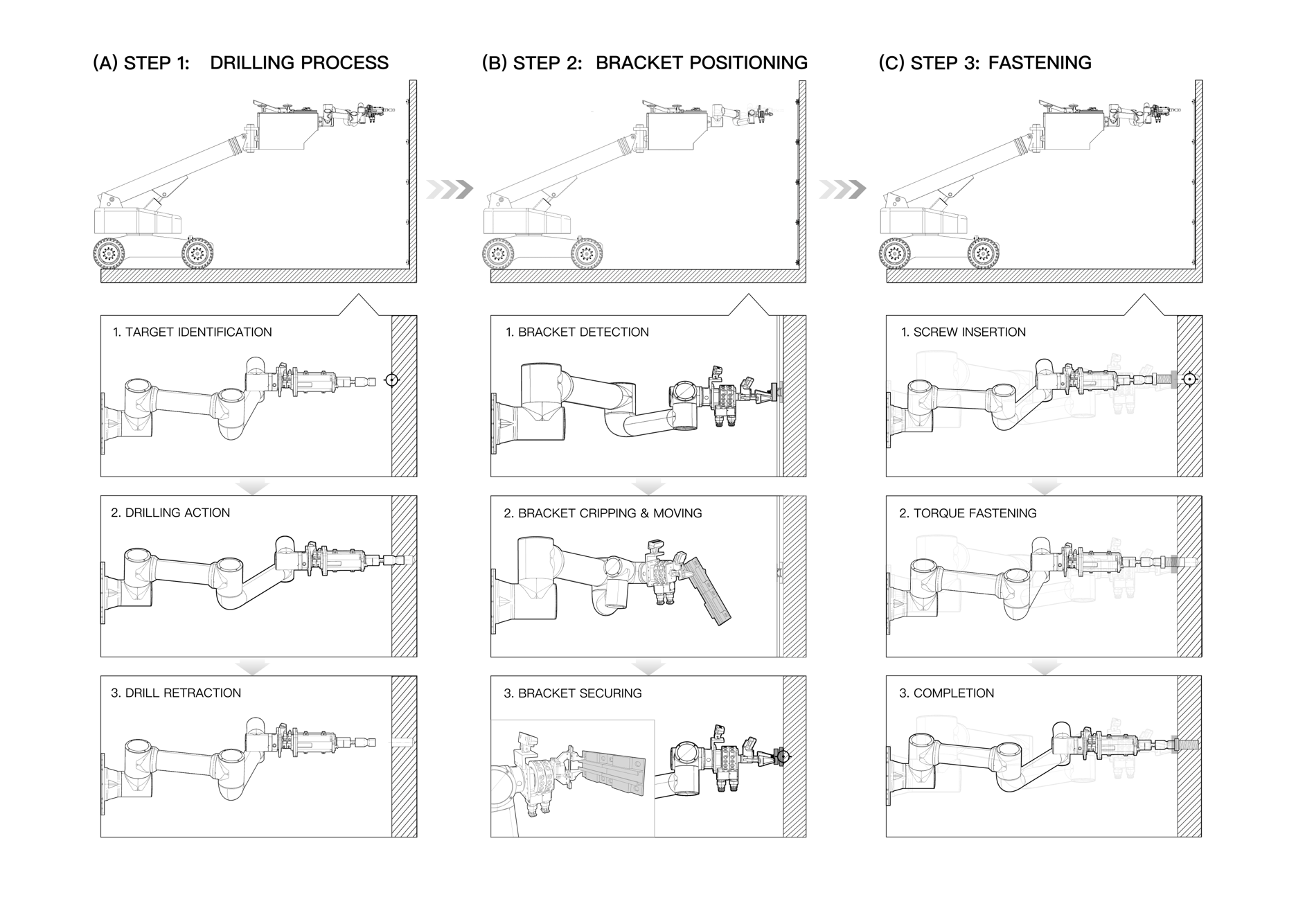}
    \caption{Workflow and system module composition for the bracket installation task.}
    \label{fig:bracket_workflow}
\end{figure}

This section presents a case study of a typical construction scenario—\textit{bracket installation}—to illustrate how the proposed semi-autonomous architecture transforms high-level user intent into coordinated dual-arm execution following the three-step pipeline described in Algorithm~\ref{algo}. The task involves drilling, bracket positioning and handling, and screw pick-and-insertion, as illustrated in Fig.~\ref{fig:bracket_workflow}. These operations require dual-arm coordination, execution in constrained workspaces, and handling multiple contact events, making this scenario an ideal benchmark for evaluating the shared-autonomy framework.

\subsection{Task Workflow Interpreted Through Algorithm~\ref{algo}}

The task is analyzed following the three-step structure of Algorithm~\ref{algo}: 
(1) intent parsing, 
(2) skill and autonomy allocation, and 
(3) execution.

\paragraph{(1) Intent Parsing (Algorithm Step 1)}
The operator provides the high-level instruction
\[
    \mathbf{T} = \text{``install the bracket at the designated façade location''}.
\]
Combined with the scene context
\[
    \mathcal{C} = \{\text{hole positions}, \text{façade plane}, \text{screw bin pose}, \text{bracket grasp points}\},
\]
the parsing module outputs task knowledge $\mathcal{K}$, a candidate sequence of three skill primitives
\[
    \mathcal{S}_c = \{\text{Drill},\ \text{bracket Handling},\ \text{Screw Handling}\},
\]
and resource assignment $R$ such as arm roles and tool utilization.

The operational workflow is then mapped to these skills:
\begin{enumerate}
    \item \textbf{Drill:} press--drill--retract cycles to prepare holes.
    \item \textbf{Bracket Handling:} transport, stabilize, and align brackets on the façade.
    \item \textbf{Screw Handling:} pick screws, insert into holes, and perform fastening.
\end{enumerate}


\paragraph{(2) Skill and Autonomy Allocation (Algorithm Step 2)}
For each skill $s_i \in \mathcal{S}_c$, the operator selects an autonomy mode $M(s_i) \in \{\text{Teleop},\ \text{Shared},\ \text{Auto}\}$.
\rev{At each control cycle, only the currently active skill $s_{i_t}$ contributes
a command. Its shared-control command is calculated according to
Eq.~\eqref{eq:active_blending}
The weights are predefined for each skill state and remain constant within
that state. They are updated only when the task state machine satisfies a
predefined transition condition.}

To clarify the role of the autonomy weights $(\alpha_i, \beta_i)$, we consider their application to key skills in the bracket-installation task:

\begin{itemize}
    \item \textbf{Full teleoperation ($\alpha_i = 1,\ \beta_i = 0$):}  
    The human operator has complete control, while the robot executes only low-level actuation. For example, during \textit{Drill}, the operator specifies both the tool position and orientation, and the robot merely follows the commanded joint motions.

\rev{
   \item \textbf{Shared control
    ($\alpha_i \neq 0,\ \beta_i \neq 0$):}
    The human and robot commands are combined using predefined state-dependent
    weights. The human may provide coarse guidance during an approach state,
    while the autonomous controller provides visual alignment or contact
    regulation. The weights do not change continuously within a state; they are
    updated only when the state machine transitions to the next execution stage.}

    \item \textbf{Full autonomy ($\alpha_i = 0,\ \beta_i = 1$):}  
    The robot independently performs perception, planning, and execution. In autonomous \textit{Drill}, it identifies the hole, aligns the end-effector, and completes the operation without human intervention.
\end{itemize}

Although full autonomy ($\alpha_i \to 0$, $\beta_i \to 1$) is desirable for minimizing human labor, façade construction involves highly unstructured and variable environments, characterized by diverse geometries, materials, and lighting conditions. Achieving full autonomy in such settings typically necessitates either task-specific reprogramming or deployment of a highly capable general-purpose AI agent. Current technical solutions, including emerging vision-language models (VLMs) and vision-language-action (VLA) models \cite{intelligence2025pi_}, are limited to relatively simple tasks (e.g., object picking or grasping) and remain insufficiently reliable for complex construction operations. These limitations reinforce the necessity of a human-in-the-loop paradigm that does not assume full autonomy as a prerequisite, but instead enables reliable task execution by systematically integrating human judgment with robotic control.

Therefore, \textbf{Shared Control} ($\alpha_i \neq 0, \ \beta_i \neq 0$) provides a practical compromise between human intuition and robotic precision. In this study, we adopt shared control for all primary manipulation tasks:
\[
    \alpha_i \neq 0, \ \beta_i \neq 0,\quad 
    \forall s_i \in \{\text{Drill},\, \text{Bracket Handling},\, \text{Screw Handling}\}.
\]

\paragraph{Discrete Subtask Allocation Ratio (SAR)}
To quantify the distribution of functional responsibility at the skill level,
we introduce a discrete metric termed the \emph{Subtask Allocation Ratio (SAR)}.
Each manipulation task is decomposed into $N$ semantically meaningful skill
primitives $\{\mathcal{S}_1, \mathcal{S}_2, \dots, \mathcal{S}_N\}$, such as target
selection, coarse approach, fine alignment, drilling, and retraction.

For each skill primitive $s_i$, a binary control label
$\delta_i \in \{0,1\}$ is assigned, where $\delta_i = 1$ indicates that the skill
is primarily human-controlled, and $\delta_i = 0$ indicates that the skill is
executed autonomously by the robot. This abstraction intentionally ignores
low-level temporal variations in control authority and focuses on the dominant
functional responsibility of each skill. The discrete SAR can be interpreted as a special case of the continuous
definition where the human control weight $\alpha_i(t)$ is binarized.

The discrete SAR is then defined as:
\begin{equation}
\mathrm{SAR} = \frac{1}{N} \sum_{i=1}^{N} \delta_i,
\label{SAR}
\end{equation}
which yields $\mathrm{SAR} \in [0,1]$. Values of SAR closer to $1$ indicate
predominantly human-driven task execution, whereas values closer to $0$
correspond to largely autonomous execution.

\rev{SAR is an aggregate measure of the proportion of human-controlled
skills and does not uniquely identify the underlying responsibility
allocation. Different allocation vectors may therefore produce the same SAR. SAR should therefore be interpreted together with the corresponding allocation vector or subtask
table.}

\begin{table}[t]
\centering
\caption{\rev{Shared-control subtask allocation for the bracket-installation case study. SAR values in this table are computed within each high-level skill (local $N{=}3$ subtasks), not across the full drilling/bracket/screw workflow.}}
\label{tab:shared_control_allocation}
\renewcommand{\arraystretch}{1.3}
\setlength{\tabcolsep}{6pt}
\begin{tabular}{p{1.9cm} p{7.2cm} p{1.2cm} p{1.3cm}}
\hline
\textbf{Task} & \textbf{Subtasks Description} & $\boldsymbol{\delta}$ & \textbf{SAR} \\
\hline
Drill 
& · Human brings the drill near the target \newline
  · Robot performs precise alignment \newline
  · Robot executes drilling
& 1 \newline
  0 \newline
  0
& 0.33 \\

\hline
Bracket Handling
& · Human guides coarse motion \newline
  · Robot stabilizes bracket \newline
  · Robot fine-tunes bracket orientation
& 1 \newline
  0 \newline
  0
& 0.33 \\

\hline
Screw Handling
& · Human provides coarse approach \newline
  · Human triggers start/stop command \newline
  · Robot aligns screw precisely and rotates
& 1 \newline
  1 \newline
  0
& 0.66 \\
\hline
\end{tabular}
\end{table}

Table~\ref{tab:shared_control_allocation} summarizes the shared-control allocation strategy for bracket installation tasks. 
Each high-level task is decomposed into a sequence of semantically meaningful subtasks, with the binary variable $\delta$ indicating whether a given subtask is executed by the human operator ($\delta=1$) or autonomously by the robot ($\delta=0$). 
This explicit skill-level allocation clarifies the functional division of responsibility between human and robot and enables a quantitative characterization of shared autonomy. 
The Subtask Allocation Ratio (SAR) is computed as the proportion of human-controlled subtasks within each task, providing a compact measure of human involvement that can be consistently compared across different manipulation skills and control modes.

\rev{
\paragraph{(3) Execution (Algorithm Step 3)}
During execution, the task state machine activates one skill at a time.
The autonomous command for the active skill is generated from the current
robot state, RGB-D perception, force/torque measurements, and contact state.
The blended command is calculated using Eq.~\eqref{eq:active_blending}, and
command conflicts are resolved using Eq.~\eqref{eq:priority_command}.

When the VR grip button activates operator override, the human command
replaces the blended command but remains subject to workspace, speed, and
force/torque limits. An E-stop or safety-interlock event immediately
suppresses all actuator commands.
}

\section{Implementation}

\begin{figure}[t]
    \centering
    \includegraphics[width=1\linewidth]{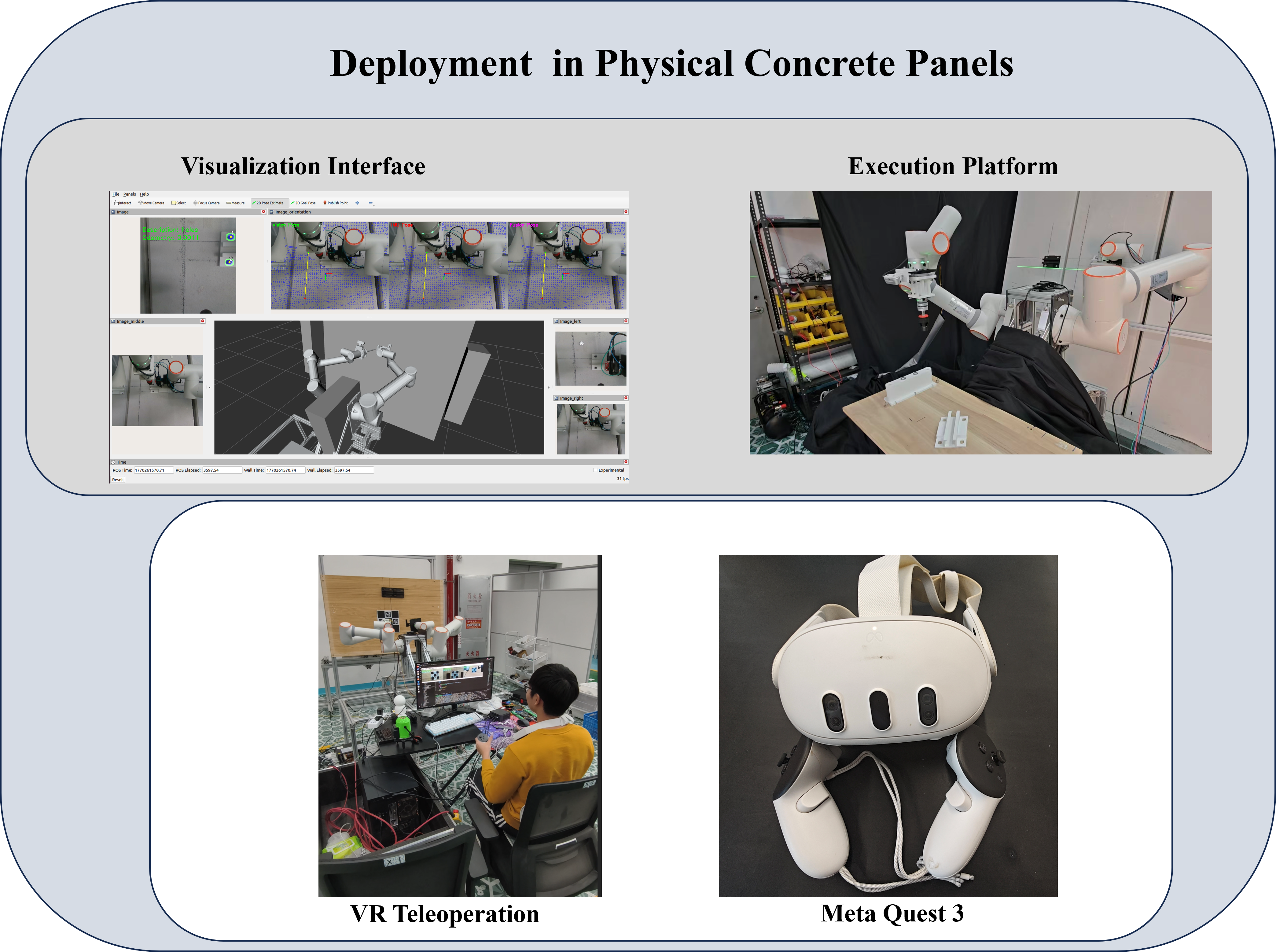}
    \caption{The deployment of the semi-autonomous dual-arm robotic system implementation.}
    \label{fig:implementation}
\end{figure}

This section details the implementation of the proposed semi-autonomous dual-arm robotic system, as illustrated in Fig.~\ref{fig:implementation}. The design integrates a VR-guided teleoperation device with an intuitive control platform, while ROS serves as the system core, coordinating motion planning and control, hole detection, facade plane estimation and pose alignment, pose mapping and workspace alignment, as well as safety management and execution. The system couples the visualization interface with the dual-arm robotic platform in a realistic construction scenario. For each aspect of the implementation, we first describe the autonomous operation, followed by the teleoperation interface, pose mapping procedures, and safety considerations.
 
\subsection{Automation Process}
\label{AutomationProcess}

This subsection presents the automated execution pipeline of the robotic system.
The overall task is decomposed into a sequence of stages, including \emph{Target Selection}, \emph{Coarse Motion}, \emph{Fine Alignment}, and \emph{Pose Correction and Insertion}, which together enable reliable autonomous operation in unstructured environments.

\begin{figure}[htp]
    \centering
    \includegraphics[width=0.5\linewidth]{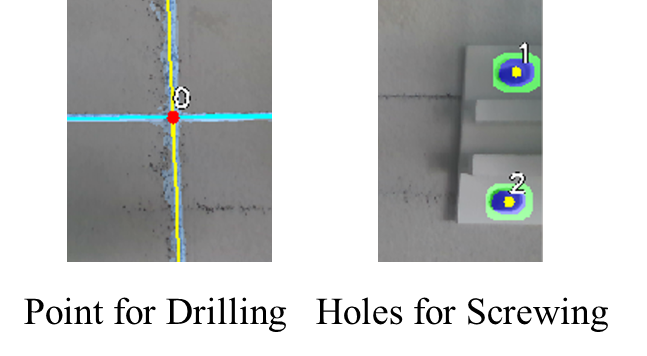}
    \caption{Predefined target selection and vision-based hole position estimation on the facade plane.}
    \label{fig:targetselection}
\end{figure}

\paragraph{Target Selection}
The target position for drilling or screw fastening is predefined on the facade plane and denoted as $\mathbf{p}_t^{\mathrm{predef}}$, as illustrated in Fig.~\ref{fig:targetselection}. 
For drilling, the target is specified by marking the intersection of two visual alignment guides on the facade surface using a construction chalk line. 
For screw fastening, the target corresponds to a previously drilled hole.

A depth camera continuously observes the workspace and estimates the actual target pose online:
\begin{equation}
\hat{\mathbf{p}}_t(t) = \arg \min_{\mathbf{p}} \mathcal{L}_{\mathrm{det}}(\mathbf{p}; \mathbf{z}(t)),
\end{equation}
where $\mathcal{L}_{\mathrm{det}}$ quantifies the discrepancy between the sensor measurements $\mathbf{z}(t)$ and the target geometric model.
The predefined target $\mathbf{p}_t^{\mathrm{predef}}$ serves as an initial nominal reference, while $\hat{\mathbf{p}}_t(t)$ provides an updated estimate of the target pose during execution.

Let $\mathbf{p}_e(t)$ denote the end-effector pose of the robot. 
The robot continuously approaches the target by minimizing the pose error
\begin{equation}
\mathbf{e}(t) = \hat{\mathbf{p}}_t(t) - \mathbf{p}_e(t),
\end{equation}
which is used to generate corrective motion commands in a closed-loop manner:
\begin{equation}
\dot{\mathbf{p}}_e(t) = \mathbf{K}\,\mathbf{e}(t),
\end{equation}
where $\mathbf{K}$ is a positive-definite gain matrix.
This feedback mechanism enables the manipulator to adapt its motion in real time, compensating for perception noise and minor target deviations to achieve stable and accurate alignment.

It should be noted that this target selection and tracking strategy exhibits limited generalization capability.
The target classes and visual appearance must be represented in the fine-tuning data, making the system sensitive to surface appearances, target types, or environmental conditions that differ substantially from those evaluated here.

\rev{\paragraph{Perception Pipeline.}
Installation targets, including drilled holes and marked target regions, are segmented using a fine-tuned YOLO26 instance-segmentation model. A detection is accepted when its confidence score exceeds $\tau_{\mathrm{det}}=0.85$ and its mask area exceeds $A_{\min}=15$ pixels. The model achieved an mIoU of $91.4\%$ on the held-out test set under the evaluated illumination and concrete-surface conditions. For a marked target, the centroid of the segmented mask defines the image-space target location. Because depth measurements inside a hole may be invalid or unstable, the segmented hole mask is dilated to construct an annular rim region, and the median valid depth within this rim is used as the target depth. The two-dimensional location and corresponding depth are back-projected through the calibrated intrinsic matrix to obtain the three-dimensional target position in the camera frame. For façade-plane estimation, the aligned depth image is converted into a point cloud. Invalid points and depths outside $[z_{\min},z_{\max}]=[0.2,3.0]$\,m are removed, after which the cloud is cropped to the façade region of interest and downsampled using a voxel size of $r_{\mathrm{voxel}}=0.02$\,m. Candidate planes are extracted by iterative RANSAC with a point-to-plane distance threshold of $d_{\mathrm{RANSAC}}=0.02$\,m and a minimum support of $N_{\min}=300$ inliers. The façade plane is selected according to inlier support and consistency between its normal and the camera viewing direction; its normal is then refined using all inliers and oriented toward the camera. Camera intrinsics and depth-to-color alignment parameters are obtained through camera calibration. The rigid camera-to-end-effector transformation is estimated offline through eye-in-hand calibration and remains fixed throughout the experiments. At runtime, the target point and façade normal are transformed from the camera frame to the robot base frame using the calibrated extrinsic transformation and robot forward kinematics. Fiducial markers are not required for façade-plane estimation. Target localization assumes that at least one drilled hole or visually distinguishable installation mark is visible and that sufficient valid depth pixels are available around it. A detection is rejected if the segmentation confidence, mask size, valid-depth ratio, or plane-inlier support fails its corresponding threshold. Under the evaluated lighting and concrete-surface conditions, the mean three-dimensional target-localization error was $12.4 \pm 3.0$\,mm and the mean façade-normal angular error was $0.61 \pm 0.056^\circ$. Sensitivity to lighting variation, partial occlusion, and dynamic workspace interference is evaluated in Sec.~\ref{sec:robustness_multi}.}

\paragraph{Coarse Motion}
Before fine alignment and contact-sensitive operations, the robot executes a coarse motion to reposition the manipulator from its current configuration to a predefined pre-alignment pose.
This stage is responsible for global, collision-free motion and does not require precise target perception or visual feedback.

Let $\mathbf{x}_e^{\mathrm{init}}$ denote the initial end-effector pose, and $\mathbf{x}_e^{\mathrm{pre}}$ denote a fixed, task-specific pre-alignment pose defined in the workspace.
The robot plans a global motion trajectory using a sampling-based planner provided by the ROS MoveIt framework:
\begin{equation}
\mathcal{T} = \text{Plan}(\mathbf{x}_e^{\mathrm{init}}, \mathbf{x}_e^{\mathrm{pre}}),
\end{equation}
where $\text{Plan}(\cdot)$ represents the RRTConnect planner operating in joint space under kinematic constraints and collision avoidance.

The planned trajectory $\mathcal{T}$ is executed in an open-loop manner, bringing the manipulator to the predefined pre-alignment pose.
Since $\mathbf{x}_e^{\mathrm{pre}}$ is fixed and independent of online perception, this stage emphasizes robustness and efficiency rather than precision.
Upon completion of the coarse motion, control is transitioned to the fine alignment module, where visual feedback is incorporated to correct residual pose errors and enable accurate insertion.

\paragraph{Fine Alignment}
To ensure normal insertion with respect to the facade surface, a local plane is estimated from the observed facade point cloud $\mathbf{P}_f = \{\mathbf{p}_i\}$:
\begin{equation}
(\mathbf{n}_f, d) = \arg \min_{\mathbf{n}, d} \sum_i \left( \mathbf{n}^\top \mathbf{p}_i + d \right)^2,
\quad \text{s.t. } \|\mathbf{n}\| = 1,
\end{equation}
where $\mathbf{n}_f$ denotes the unit normal vector of the facade plane.
Based on this estimate, the desired end-effector orientation is defined as
\begin{equation}
\mathbf{R}_r^* = \text{align}(\mathbf{z}_e, \mathbf{n}_f),
\end{equation}
where $\mathbf{z}_e$ is the current approach axis of the end-effector, and $\text{align}(\cdot)$ computes the rotation that aligns $\mathbf{z}_e$ with the surface normal.

\paragraph{Pose Correction and Insertion}
The desired target pose for insertion is given by
\begin{equation}
\mathbf{x}_e^* =
\begin{bmatrix}
\hat{\mathbf{p}}_t \\
\mathbf{R}_r^*
\end{bmatrix},
\end{equation}
where $\hat{\mathbf{p}}_t$ is the online estimated target position.
Let $\mathbf{x}_e(t) = [\mathbf{p}_e(t), \mathbf{R}_e(t)]$ denote the current end-effector pose.
The robot performs closed-loop pose correction by minimizing the position and orientation errors
\begin{equation}
\mathbf{e}_x(t) =
\begin{bmatrix}
\hat{\mathbf{p}}_t - \mathbf{p}_e(t) \\
\text{Log}\!\left(\mathbf{R}_e(t)^\top \mathbf{R}_r^*\right)
\end{bmatrix},
\end{equation}
which are used to generate corrective motion commands during the approach phase.

Once the pose error falls below a predefined threshold, the robot executes the insertion motion along the estimated surface normal:
\begin{equation}
\mathbf{p}_e(t+\Delta t) = \mathbf{p}_e(t) + v \, \mathbf{n}_f \, \Delta t,
\end{equation}
where $v$ is a constant insertion velocity.
This sequential alignment and insertion strategy enables accurate and stable execution while reducing the risk of jamming or surface damage.

\rev{\paragraph{Contact Control}
A six-axis force/torque sensor is mounted between the robot end effector
and the task tool to measure the interaction force and torque in real time.
During contact-rich execution, motion along the tool axis is regulated by
force control, while the remaining translational and rotational degrees of
freedom are maintained at the aligned target pose. This provides compliance
in the contact direction and avoids purely position-controlled penetration
into the workpiece.

During rule-based autonomous drilling, the robot applies a constant axial
feed force of 30\,N, while the depth camera monitors the feed depth. A depth
of 50\,mm is used as the nominal controller setpoint; reaching this setpoint
terminates the autonomous feed motion and triggers tool retraction. The
drilling-motor current is monitored simultaneously for jamming detection.
A jam is identified when the current remains above 5\,A for more than 2\,s,
after which the controller immediately stops the axial feed motion and
retracts the drill.

During rule-based autonomous screw fastening, insertion depth and motor
torque are monitored throughout the operation. An insertion depth of 50\,mm
is used as the nominal controller target. A premature torque rise before
stable screw engagement is treated as jamming, causing the fastening motion
to stop and the tool to retract.

These controller targets define the desired termination conditions for the
rule-based autonomous skills. They are distinct from the binary
task-success criteria used in the cross-mode experiments, in which achieved
depth, position error, orientation error, and perpendicularity are analyzed
as continuous measures of execution quality.}

\subsection{VR-Guided Teleoperation}

\begin{figure}[t]
    \centering
    \includegraphics[width=0.85\linewidth]{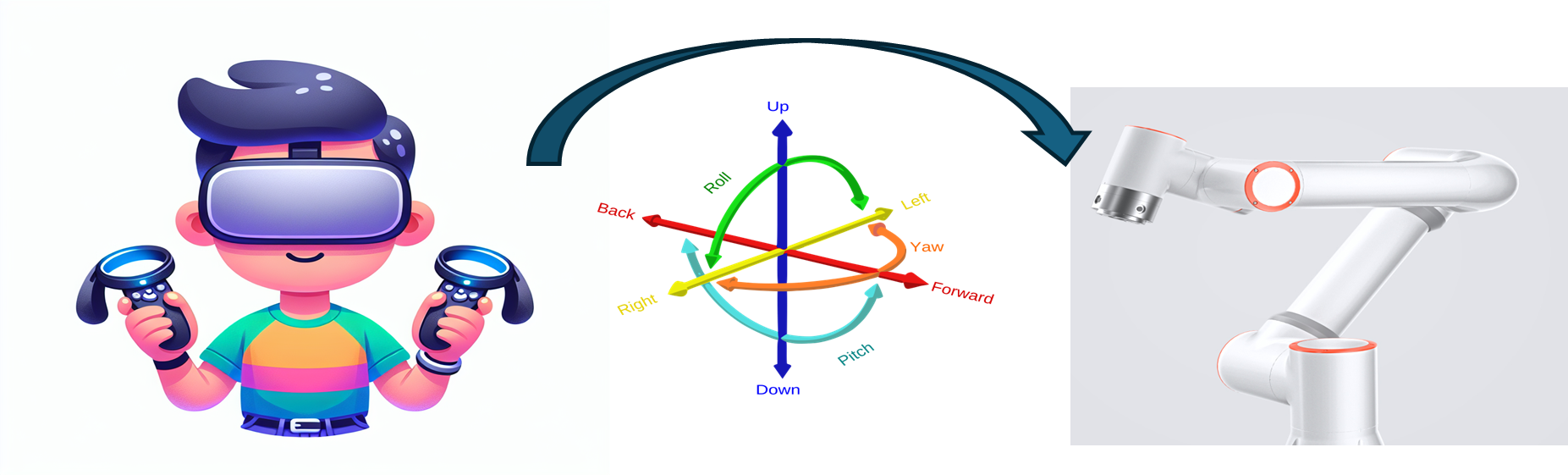}
    \caption{Schematic of the VR teleoperation interface. Human motion is mapped to dual-arm robotic motion via the controllers.}
    \label{fig:vr_interface}
\end{figure}

To complement autonomous execution, a VR-based teleoperation interface allows intuitive human control of the dual-arm system. As shown in Fig.~\ref{fig:vr_interface}, the operator wears a VR headset and manipulates the arms via handheld controllers. The human end-effector pose is
\begin{equation}
\mathbf{x}_h(t) = 
\begin{bmatrix} \mathbf{p}_h(t) \\ \mathbf{R}_h(t) \end{bmatrix},
\end{equation}
which is mapped in real time to the robot's end-effector pose:
\begin{equation}
\mathbf{x}_r(t) = \mathbf{T}_{hr} \cdot \mathbf{x}_h(t),
\end{equation}
where $\mathbf{T}_{hr} \in SE(3)$ aligns human and robot reference frames.

\begin{figure}[t]
    \centering
    \includegraphics[width=0.5\linewidth]{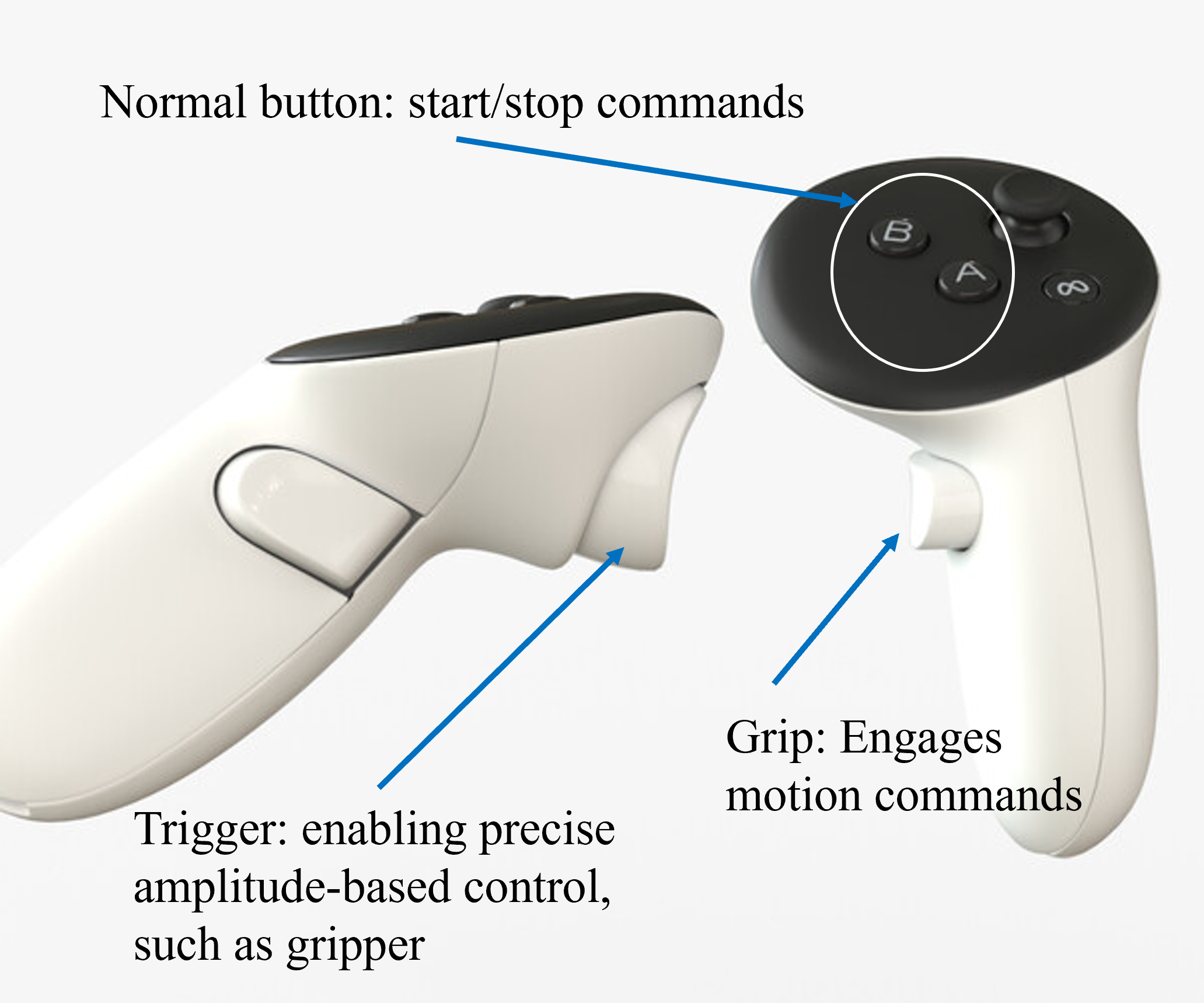}
    \caption{Mapping of VR controller inputs to robot skill primitives.}
    \label{fig:Controller}
\end{figure}

\paragraph{VR Trigger Mechanism} 
The mapping of operator inputs from the VR controllers to robotic skill primitives is illustrated in Fig.~\ref{fig:Controller} and summarized as follows:

\begin{itemize}
    \item \textbf{Trigger:} Modulates the gripper opening and closing, enabling precise amplitude-based control for fine manipulation.
    \item \textbf{Normal buttons:} including the A/B/X/Y button. Serve as start/stop commands for task-specific operations, such as drilling or screwing. They can also initiate automatic alignment procedures and toggle semi-autonomous skill execution.
    \item \textbf{Grip button:} Engages motion commands; the corresponding robot arm responds only while the button is actively pressed, allowing real-time operator intervention.
\end{itemize}

\paragraph{Pose Mapping and Workspace Alignment}

To adapt human motion to the robot's workspace, we apply relative scaling:
\begin{equation}
\mathbf{p}_r(t) = s (\mathbf{p}_h(t) - \mathbf{p}_h^0) + \mathbf{p}_r^0,
\end{equation}
with $s$ as the workspace scaling factor and $\mathbf{p}_h^0, \mathbf{p}_r^0$ the initial reference positions. Orientation is aligned via
\begin{equation}
\mathbf{R}_r(t) = \mathbf{R}_{hr} \mathbf{R}_h(t),
\end{equation}
ensuring consistent human-to-robot mapping.

This explicit mapping relieves the operator from low-level joint or force control, allowing focus on high-level task execution and strategic decision-making.

\subsection{ROS-Based Safety and Execution}

Throughout task execution, the operator retains full supervisory authority via the ROS-based control framework, which seamlessly integrates autonomous operation, teleoperation, sensor processing, and obstacle avoidance. Through ROS, the operator can override autonomous commands, assume manual control, pause tasks, adjust the robot’s configuration, or activate the emergency stop (E-stop) to immediately halt all motion and lock the actuators. Sensor signals from the perception system are continuously processed within ROS to inform motion planning, obstacle detection, and dynamic path adjustment, \rev{supporting safer navigation} in complex, unstructured environments.  

In semi-autonomous operation, where control is shared between human and robot, these ROS-managed mechanisms provide continuous oversight and immediate intervention capabilities, preserving transparency, controllability, and overall system safety. By integrating supervisory control, sensor processing, and obstacle avoidance within the ROS ecosystem, the system allows the operator to guide, correct, or halt robot actions in response to environmental uncertainties or unexpected events, thereby \rev{reducing the risk of} collision and unsafe motion under real-world conditions.

\rev{\paragraph{Safety Analysis.}
MoveIt collision checking is enabled for both arms and the façade fixture. Cartesian speed is limited to 0.15\,m/s in teleoperation and 0.08\,m/s during contact. Workspace soft limits are enforced in software; hardware E-stop latency was measured at approximately 120\,ms. During all recorded trials, no unintended collisions occurred; two semi-autonomous drilling trials triggered force-limit retraction and operator recovery. VR-to-robot command latency averaged 45\,ms. These observations support safe supervised operation in the tested mock-up but do not constitute a formal safety certification.}

\rev{\subsection{Hardware and Software Stack}}
\rev{Table~\ref{tab:hw_sw} summarizes the hardware and software configuration used for reproducibility.}

\begin{table}[t]
\centering
\caption{\rev{Hardware and software stack.}}
\label{tab:hw_sw}
\footnotesize
\resizebox{\textwidth}{!}{%
\begin{tabular}{ll}
\toprule
\textbf{Component} & \textbf{Specification} \\
\midrule
Robot arms & 2$\times$ FAIRINO FR5 (6-DoF), controller update 125\,Hz \\
End-effectors & Parallel gripper, drill head, electric screwdriver, suction gripper \\
Force/torque sensing & Six-axis sensor mounted between the end effector and task tool \\
RGB-D camera & Intel RealSense D435, 30\,Hz \\
VR device & Meta Quest 3 with 6-DoF controllers \\
Workstation & Intel i9 CPU, 32\,GB RAM, NVIDIA RTX 4080 \\
Middleware & ROS Noetic, MoveIt, RRTConnect planner, KDL IK \\
Control loop & Cartesian command rate 50\,Hz; measured VR-to-robot latency 45\,ms \\
\bottomrule
\end{tabular}%
}
\end{table}

\rev{\subsection{Task State Machines}}
\rev{Tables~\ref{tab:state_drill}--\ref{tab:state_screw} summarize the
implementation-level state machines for drilling, bracket pick-and-place,
and dual-arm screw fastening. The tables report the principal execution
states, sensor inputs, transition conditions, operator-override paths, and
representative failure states. Quantitative controller limits and binary
task-completion criteria are reported separately in the corresponding
control-implementation and experimental-protocol sections.}

\begin{table}[t]
\centering
\caption{\rev{Drilling task state machine.}}
\label{tab:state_drill}
\footnotesize
\resizebox{\textwidth}{!}{%
\begin{tabular}{lllll}
\toprule
\textbf{State} & \textbf{Sensors} & \textbf{Transition} & \textbf{Override} & \textbf{Failure} \\
\midrule
Target Select & RGB-D & Marker/hole found & Grip pressed & No detection \\
Coarse Move & Joint encoders & Reach pre-pose & Grip pressed & Planner fail \\
Fine Align & RGB-D, wrench & Alignment accepted & Grip pressed & Align timeout \\
Drill & Force, current, depth &
\makecell[l]{Nominal depth reached\\or drilling terminated} &
E-stop & Jamming detected \\
Retract & Joint encoders & Safe retract done & Grip pressed & Collision risk \\
\bottomrule
\end{tabular}%
}
\end{table}

\begin{table}[t]
\centering
\caption{\rev{Bracket pick-and-place state machine.}}
\label{tab:state_bracket}
\footnotesize
\resizebox{\textwidth}{!}{%
\begin{tabular}{lllll}
\toprule
\textbf{State} & \textbf{Sensors} & \textbf{Transition} & \textbf{Override} & \textbf{Failure} \\
\midrule
Pick & Gripper state & Grasp confirmed & Grip pressed & Slip \\
Flip Arm & Joint encoders & Reorient done & Grip pressed & Singularity \\
Target Select & RGB-D & Target valid & Grip pressed & No target \\
Coarse/Place & RGB-D, encoders & Placement pose & Grip pressed & Reach error \\
Fine Align & RGB-D & Alignment accepted & Grip pressed & Timeout \\
\bottomrule
\end{tabular}%
}
\end{table}

\begin{table}[t]
\centering
\caption{\rev{Dual-arm screw fastening state machine.}}
\label{tab:state_screw}
\footnotesize
\resizebox{\textwidth}{!}{%
\begin{tabular}{lllll}
\toprule
\textbf{State} & \textbf{Sensors} & \textbf{Transition} & \textbf{Override} & \textbf{Failure} \\
\midrule
Screw Pick & Gripper & Grasp confirmed & Grip pressed & Drop \\
Flip Arm & Encoders & Reorient done & Grip pressed & Reach fail \\
Hole ID & RGB-D & Hole detected & Grip pressed & False detect \\
Coarse Approach & RGB-D & Near hole & Grip pressed & Misalignment \\
Fine Align & RGB-D, wrench & Alignment accepted & Grip pressed & Timeout \\
Fasten & F/T, depth &
\makecell[l]{Nominal fastening\\targets reached} &
E-stop & \makecell[l]{Premature torque rise\\or jamming} \\
Right-arm Hold & Gripper force & Stable hold & Grip pressed & Slip \\
\bottomrule
\end{tabular}%
}
\end{table}

\paragraph{Summary}
The system follows a modular pipeline comprising VR-based input acquisition, pose processing, command generation, and robot execution. Hand positions $\mathbf{p}_h(t)$ and orientations $\mathbf{R}_h(t)$ from the VR controllers are mapped into the robot workspace through scaling and alignment, then converted into Cartesian or joint-space commands $\mathbf{u}(t)$. These commands are executed via inverse kinematics and coordinated dual-arm control under safety constraints. Overall, the architecture combines human intuition with robotic precision, providing a compact, safe, and adaptable framework for complex construction tasks.


\section{Experiments and Results}

\subsection{Experimental Setup}

To evaluate the proposed semi-autonomous dual-arm robotic system, experiments were conducted
in a mock construction environment designed to capture key characteristics of real-world
building sites, including vertical façades, limited viewpoints, and frequent task
interruptions. Three representative construction operations were selected: drilling, bracket pick-and-place, and screw fastening, which together form a typical façade installation workflow.

Three control paradigms were compared:

\begin{enumerate}
    \item \textbf{Full VR Teleoperation (TeleOp):}  
    The human operator directly controls both robot arms in full 6-DoF via a VR interface, encompassing target localization, approach, alignment, actuation, and task termination.

    \item \textbf{Semi-autonomous Shared Control (Semi-Auto):}  
    The operator provides high-level guidance, including target selection and coarse motion commands, while the robot autonomously executes fine alignment, contact-consistent manipulation, and local motion refinement.

    \item \textbf{\rev{Rule-Based Autonomous Control (Auto):}}  
    \rev{The robot executes the task without human intervention once triggered, using predefined targets, preprogrammed waypoints, and a fixed rule-based execution pipeline. This baseline should not be interpreted as general-purpose or learning-based autonomy.}
\end{enumerate}

\rev{It is important to clarify that the Auto mode evaluated in this study represents conventional rule-based automation under predefined laboratory conditions. Once triggered by the experimenter, Auto runs without operator teleoperation or mid-trial correction: it uses pre-programmed waypoints and calibrated perception pipelines, without online replanning or adaptation to environmental changes. Consequently, lighting degradation, occlusion, or dynamic workspace interference may lead to task failure. Robustness under these disturbances is evaluated for the drilling task in Sec.~\ref{sec:robustness_drilling}.}

\begin{figure}[htbp]
    \centering
    \includegraphics[width=0.75\linewidth]{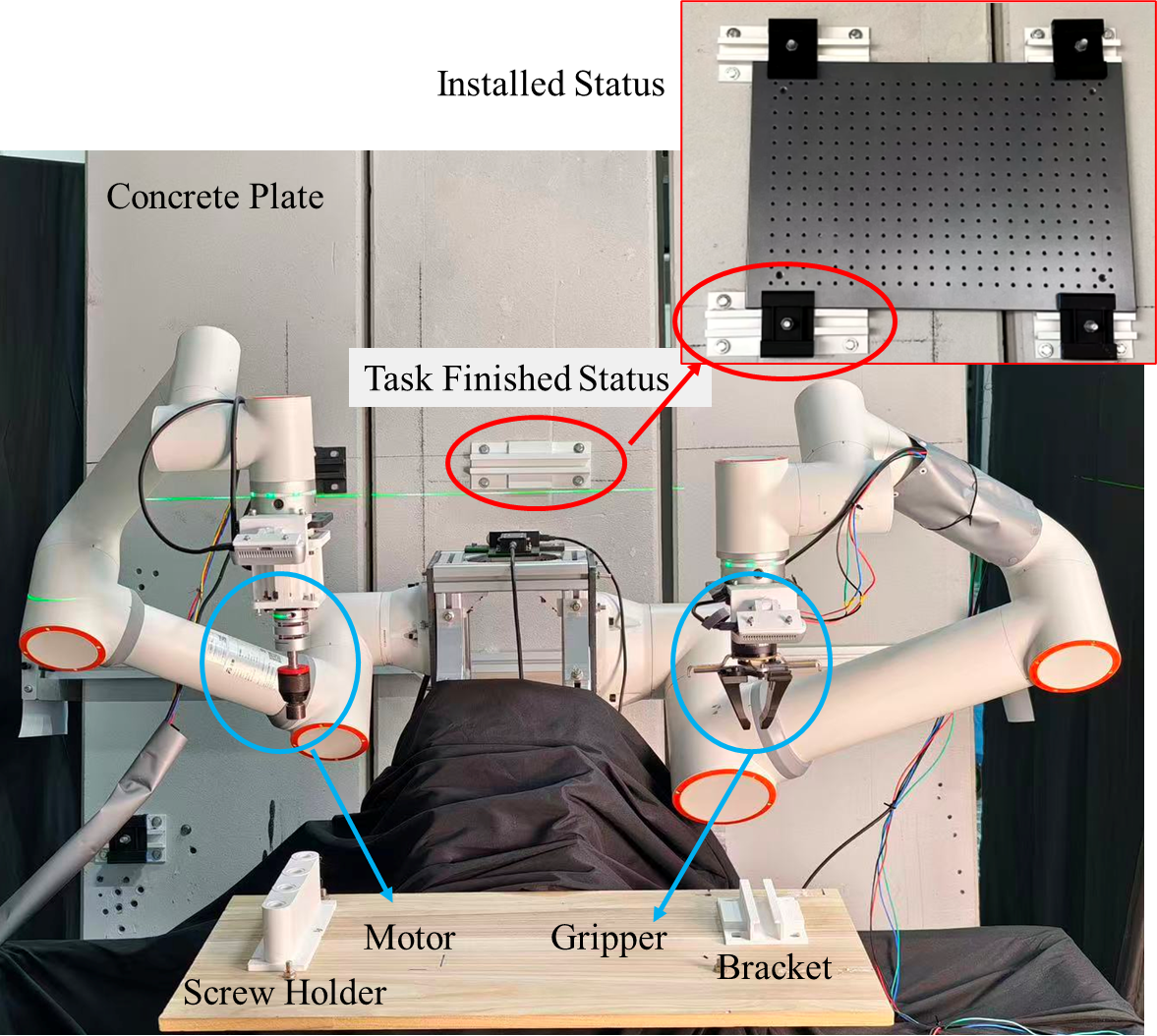}
    \caption{Experimental setup for evaluating the dual-arm robotic system in the bracket assembly task.}
    \label{fig:experimentsetup}
\end{figure}

\begin{figure}[htbp]
    \centering
    \includegraphics[width=1\linewidth]{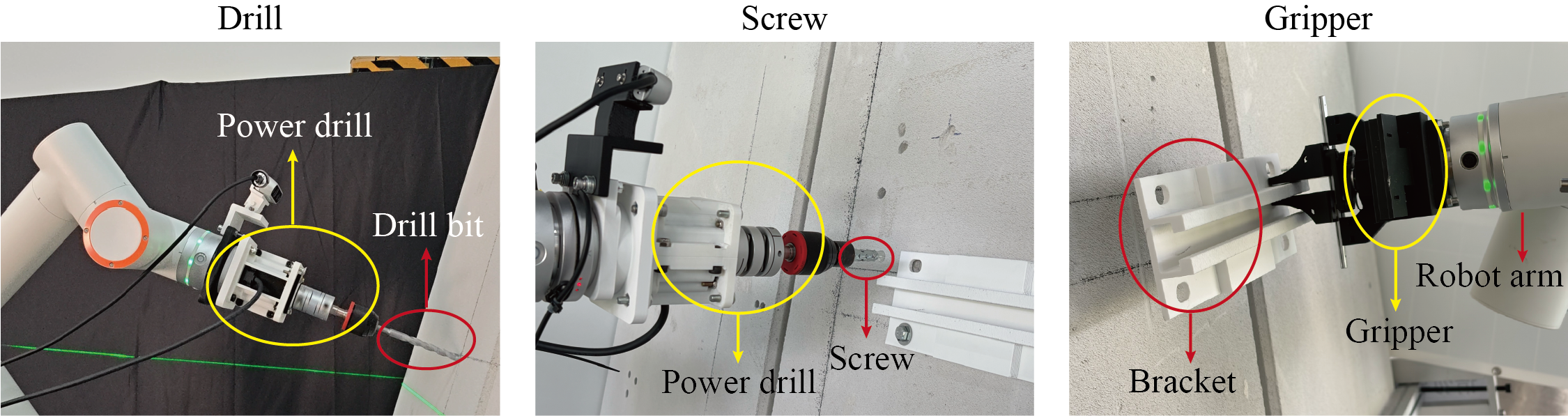}
    \caption{End-effector configurations used for drilling, screw fastening, and bracket manipulation tasks.}
    \label{fig:endeff}
\end{figure}

The experimental setup and overall system architecture are illustrated in Fig.~\ref{fig:experimentsetup}, 
while the end-effector configurations are shown in Fig.~\ref{fig:endeff}. 
The dual-arm system consists of one arm equipped with a parallel gripper for object handling, 
and a second arm driven by a motorized tool interface. 
Depending on the task, the tool arm is fitted with either a drilling head for the drilling task or a screw-driving head for the screw fastening task. 
Screws are preloaded in a holder, and brackets are placed on a dedicated working plate. 
A concrete plate is used as the target surface for all contact-rich manipulation tasks. After task completion, the bracket is securely mounted on the concrete plate, with four screws fastened through the designated installation holes. 
The installed bracket is thus ready for subsequent mounting or functional use on a vertical wall surface.

To increase task difficulty and evaluate operator difficulty under constrained viewpoints, 
the working plate is positioned on the opposite side of the concrete plate. 
This configuration requires the robot to perform a \emph{Flip the Arm} maneuver to access the target, 
which is particularly challenging under pure teleoperation due to limited perception and complex kinematic coupling.

\subsection{Experimental Protocol}
\label{sec:experimental_protocol}

\rev{Ten novice operators with no prior VR teleoperation experience participated in the human-operated conditions (TeleOp and Semi-Auto). Ethics approval was obtained from the institutional review board, and written informed consent was collected before participation. Each participant received a 15-minute safety briefing, a 20-minute VR familiarization session, and three excluded practice trials per task under the human-operated modes. The order of TeleOp and Semi-Auto was counterbalanced across participants by random assignment to two order groups (TeleOp-first vs.\ Semi-Auto-first). Each participant completed 10 recorded TeleOp trials and 10 recorded Semi-Auto trials per task after practice, yielding \(10 \times 10{=}100\) participant trials per task for each of TeleOp and Semi-Auto. Rest periods of at least 2 minutes were provided between blocks. Trial-order trends were inspected descriptively; no monotonic performance gain across recorded trial index was evident that would reverse the mode ranking.}

\rev{Rule-based Auto does \emph{not} require a human operator during execution. For each task, Auto was evaluated in \(N{=}100\) triggered repetitions under the same fixture and calibrated perception setting as the human-operated blocks, without VR teleoperation or mid-trial human correction. Auto counts therefore reflect autonomous pipeline repeats rather than participant sessions.}

\rev{Table~\ref{tab:master_decomp} provides a master task-decomposition table defining subtasks, human/robot responsibility, control mode, SAR, and measured outputs for each evaluated task. SAR is always computed using the task-specific subtask count $N$ listed in the table.}

\begin{table}[t]
\centering
\caption{\rev{Master task decomposition, responsibility allocation, and SAR definitions.}}
\label{tab:master_decomp}
\footnotesize
\renewcommand{\arraystretch}{1.1}
\resizebox{\textwidth}{!}{%
\begin{tabular}{llllll}
\toprule
\textbf{Task} & \textbf{Subtasks ($N$)} & \textbf{Human / Robot responsibility (Semi-Auto)} & \textbf{Mode} & \textbf{SAR} & \textbf{Outputs} \\
\midrule
Drilling & 5 skills & Human: select, coarse, retract; Robot: align, drill & Semi-Auto & 0.6 & Success, depth, pos., perp., time \\
Bracket & 6 skills & Human: pick, coarse, select, place; Robot: flip, align & Semi-Auto & 0.67 & Success, pos., ori., time \\
Screw & 7 skills & Human: pick, coarse, hole ID, right-arm hold; Robot: flip, align, fasten & Semi-Auto & 0.57 & Success, insertion, time \\
\bottomrule
\end{tabular}%
}
\end{table}

\rev{\paragraph{Task-Success Criteria and Measurement Methods}
Binary task success was evaluated separately from continuous execution
quality. For drilling, a trial was considered successful when the drill bit
penetrated the wall and produced a measurable hole, provided that no tool
jamming, safety-stop activation, or task timeout occurred. For bracket
pick-and-place, a trial was considered successful when the bracket was
released at the intended installation region and remained stably placed
without slipping for 10\,s. For screw fastening, a trial was considered
successful when the screw entered the target hole or connection and remained
engaged after screwdriver retraction, provided that no screw dropping,
jamming, safety-stop activation, or task timeout occurred. A trial not
completed within 300\,s was classified as a timeout failure.

The prescribed drilling depth of 50\,mm and screw-insertion depth of 50\,mm
were treated as nominal controller targets rather than mandatory binary
success thresholds. Position, orientation, and perpendicularity errors were
measured relative to ideal references of 0\,mm or $0^\circ$, as applicable.
The bracket-position and orientation values of 25\,mm and $15^\circ$ were
treated only as reference tolerances for interpreting placement quality,
not as binary success thresholds. These continuous measures characterize
how accurately and completely a task was executed, whereas the success rate
indicates whether the basic physical operation was completed.

During execution, drilling depth was monitored using the depth camera,
screw-insertion depth was monitored by the fastening controller, and
tightening torque was measured by the end-effector-mounted six-axis
force/torque sensor. For outcome reporting, geometric measures were also
verified manually after each trial. Positional error was measured relative
to the marked target using a ruler or digital caliper, orientation error and
drilling perpendicularity were measured using an angle gauge, and drilling
and screw-insertion depths were verified using a digital caliper. Task time
was recorded from task initiation to success or timeout.}

\rev{\paragraph{Statistical Reporting.}
The experimental outcomes are reported descriptively. Success rate is
reported as the pooled proportion of trials satisfying the binary
task-completion criteria. For both the nominal three-mode comparison and the
drilling robustness experiment, continuous outcomes are summarized as trial-level
mean \(\pm\) sample standard deviation over successful trials. The dedicated
SAR sweep retains the observed mean values because trial-level dispersion was
not available for that protocol. Raw NASA-TLX outcomes are summarized as
mean \(\pm\) sample standard deviation for each drilling-mode condition.}

We evaluate the proposed shared-control framework on three representative manipulation tasks: a \textit{Drilling Task}, a \textit{Bracket Pick-and-Place Task}, and a \textit{Dual-Arm Screw Fastening Task}.

\rev{For low-success TeleOp conditions, the reported time and error metrics are calculated only over successful trials and are therefore not representative of all attempts. Success rate and the qualitative failure analysis should be considered together with these conditional continuous metrics. Tables~\ref{tab:drilling_protocol}--\ref{tab:screw_protocol} report the three-mode comparison under nominal conditions.}

\begin{figure}[htbp]
    \centering
    \includegraphics[width=0.75\linewidth]{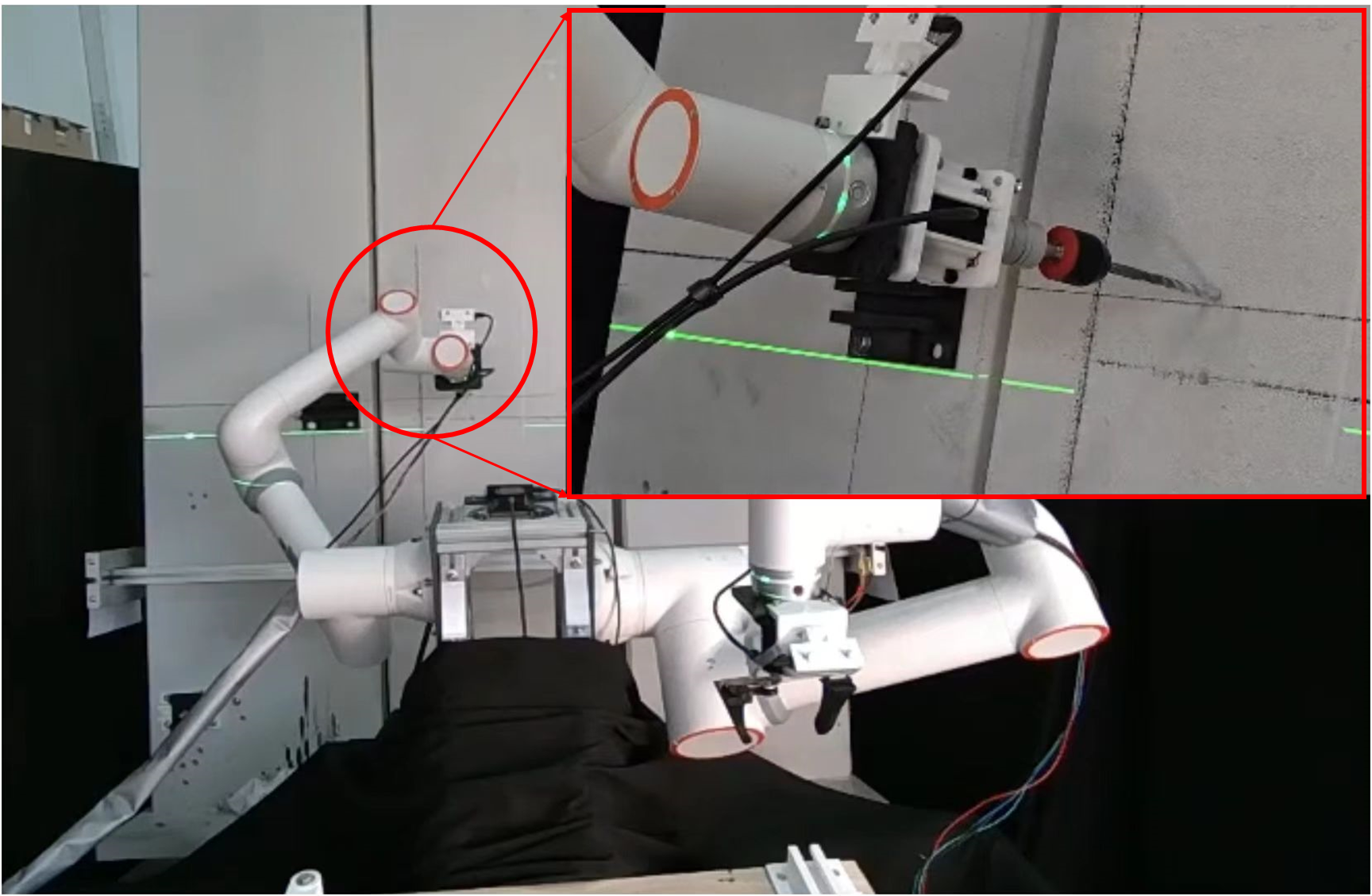}
    \caption{Drilling Task.}
    \label{fig:DrillingTask}
\end{figure}

\paragraph{Task 1: Drilling Task.}
The drilling task, shown in Fig.\ref{fig:DrillingTask}, is decomposed into five sub-tasks: \emph{Target Selection}, \emph{Coarse Approach}, \emph{Fine Alignment}, \emph{Drilling}, and \emph{Retraction}. Under semi-autonomous control, the human operator is responsible for \emph{Target Selection}, \emph{Coarse Approach}, and \emph{Retraction}, while the robot autonomously performs \emph{Fine Alignment} and \emph{Drilling}, yielding a Subtask Allocation Ratio (SAR) of $3/5 = 60\%$.

\rev{A trial is considered successful when the drill penetrates the wall and produces a measurable hole without jamming, safety-stop activation, or timeout. The achieved depth, target-position error, and perpendicularity are evaluated separately as continuous measures of drilling quality.}
Figure~\ref{fig:drill_compare} compares rule-based autonomous drilling with teleoperated execution. 
The rule-based Auto mode achieves more consistent and accurate drilling performance, whereas teleoperation frequently results in misaligned holes or insufficient drilling clearance due to human-induced positioning errors.

\rev{
\begin{figure}[h]
    \centering
    \includegraphics[width=0.75\linewidth]{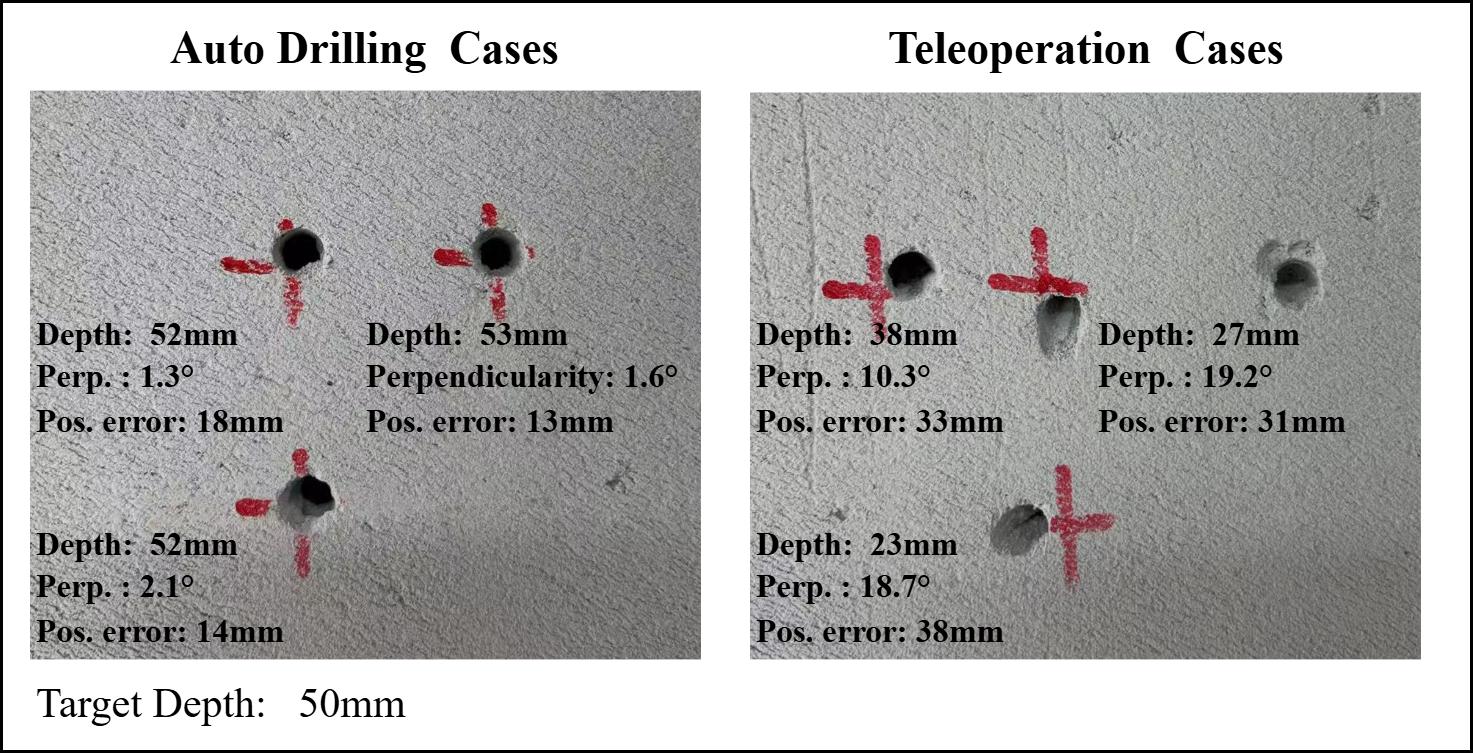}
    \caption{Comparison of autonomous and teleoperated drilling results. Autonomous execution produces more accurate and consistent holes, while teleoperation often leads to misalignment or insufficient clearance.}
    \label{fig:drill_compare}
\end{figure}
}

{\color{blue}
\begin{table}[htbp]
\centering
\caption{Performance of the three control modes in the nominal drilling experiment. TeleOp and Semi-Auto each include 100 trials nested within 10 participants; Auto includes 100 unmanned pipeline repetitions. Continuous metrics are reported as trial-level mean $\pm$ sample standard deviation over successful trials; TeleOp metrics are based on 53 successful trials.}
\label{tab:drilling_protocol}
\resizebox{\linewidth}{!}{%
\begin{tabular}{lccccc}
\hline
Mode & Success (\%) & Depth (mm) & Pos. error (mm) & \makecell{Perp.\\($^\circ$)} & Time (s) \\
\hline
TeleOp    & 53 & $31.2 \pm 7.4$ & $36.4 \pm 9.2$ & $15.0 \pm 4.8$ & $92.5 \pm 18.6$ \\
Semi-Auto & 82 & $50.6 \pm 2.3$ & $23.3 \pm 4.1$ & $2.2 \pm 0.7$ & $38.7 \pm 5.8$ \\
Auto      & 96 & $52.1 \pm 1.7$ & $15.0 \pm 2.5$ & $1.8 \pm 0.5$ & $15.0 \pm 2.0$ \\
\hline
\end{tabular}%
}
\end{table}
}

\rev{Semi-autonomous control substantially improves success rate and geometric accuracy compared to pure teleoperation. Rule-based Auto achieves the highest efficiency and perpendicularity under predefined conditions, but its dependence on calibrated targets limits expected generalization.}

\begin{figure}[ht]
    \centering
    \includegraphics[width=0.75\linewidth]{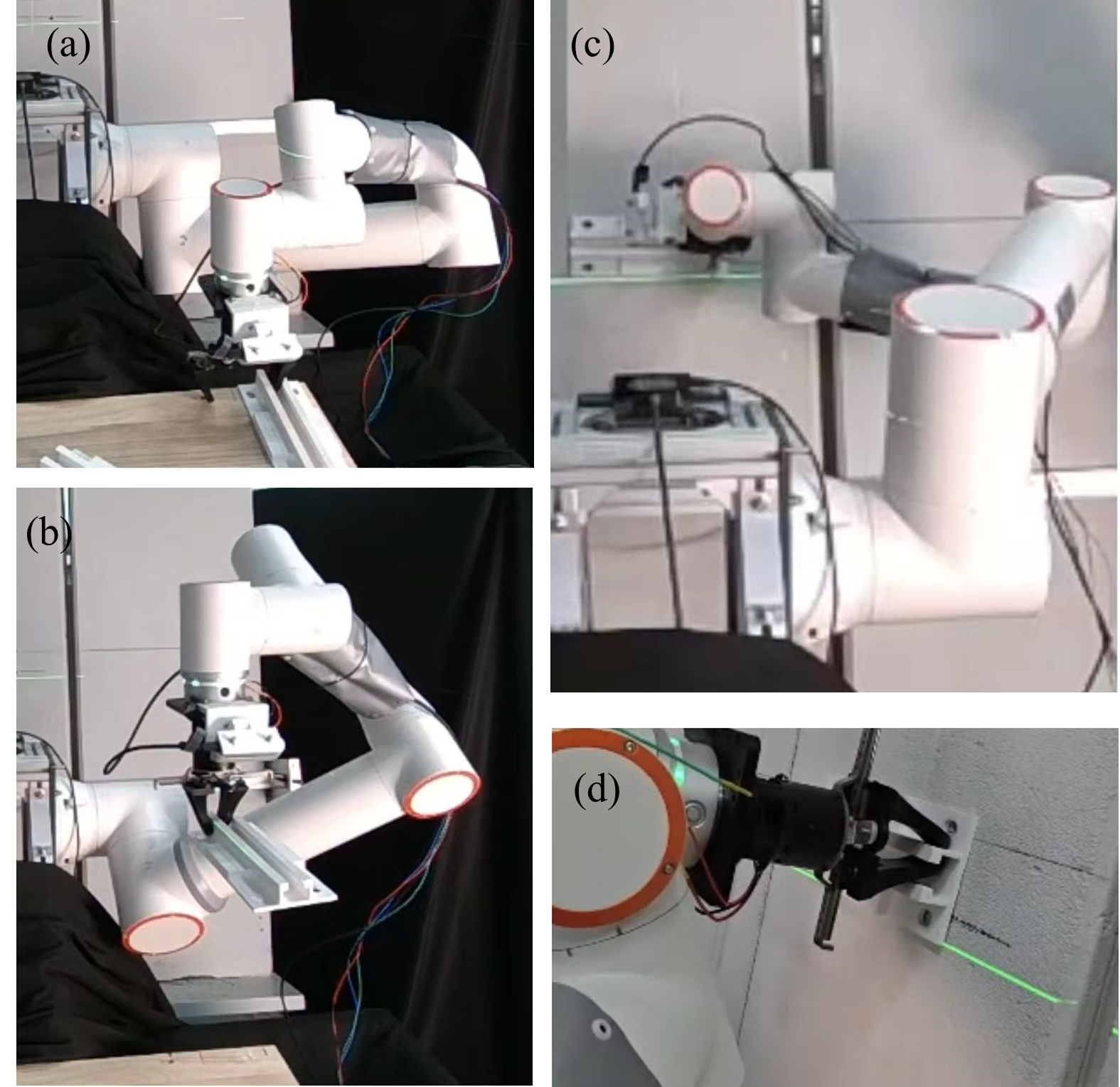}
    \caption{Bracket pick-and-place task. The figure illustrates representative stages of the task, including 
    (a) gripper approaching the bracket, 
    (b) bracket picking, 
    (c) flipping the arm, and 
    (d) final placement against the wall.}
    \label{fig:brasket_case}
\end{figure}

\paragraph{Task 2: Bracket Pick-and-Place Task.}
As illustrated in Fig.~\ref{fig:brasket_case}, this task is decomposed into six sub-tasks: 
\emph{Bracket Pick}, \emph{Flip the Arm}, \emph{Target Selection}, \emph{Coarse Approach}, \emph{Fine Alignment},  and \emph{Placement}. 
The human operator performs \emph{Bracket Pick}, \emph{Coarse Approach}, \emph{Target Selection}, and \emph{Placement}, 
while the robot autonomously executes \emph{Flip the Arm} and \emph{Fine Alignment}, resulting in a Subtask Allocation Ratio (SAR) of $4/6 \approx 0.67$.

\rev{A trial is considered successful when the bracket is released at the
intended installation region and remains stably placed without slipping for
10\,s. Position and orientation errors are evaluated separately as
continuous measures of placement quality.}

{\color{blue}
\begin{table}[htbp]
\centering
\caption{Performance of the three control modes in the nominal bracket pick-and-place experiment. TeleOp and Semi-Auto each include 100 trials nested within 10 participants; Auto includes 100 unmanned pipeline repetitions. Continuous metrics are reported as trial-level mean $\pm$ sample standard deviation over successful trials; TeleOp metrics are based on 4 successful trials.}
\label{tab:bracket_protocol}
\resizebox{\linewidth}{!}{%
\begin{tabular}{lcccc}
\hline
Mode & Success (\%) & Pos. Error (mm) & Ori. Error ($^\circ$) & Time (s) \\
\hline
TeleOp    & 4 & $18.6 \pm 6.9$ & $11.8 \pm 4.2$ & $198.6 \pm 38.5$ \\
Semi-Auto & 81 & $3.1 \pm 1.0$ & $2.4 \pm 0.8$ & $74.3 \pm 10.5$ \\
Auto      & 97 & $2.0 \pm 0.6$ & $1.6 \pm 0.5$ & $41.2 \pm 6.2$ \\
\hline
\end{tabular}
}
\end{table}
}

\rev{Pure teleoperation fails in most cases due to the difficulty of large-angle manipulator reorientation. Semi-autonomous control achieves substantially higher success rates than TeleOp, while rule-based Auto remains best under predefined target configurations.}

\begin{figure}[htbp]
    \centering
    \includegraphics[width=1\linewidth]{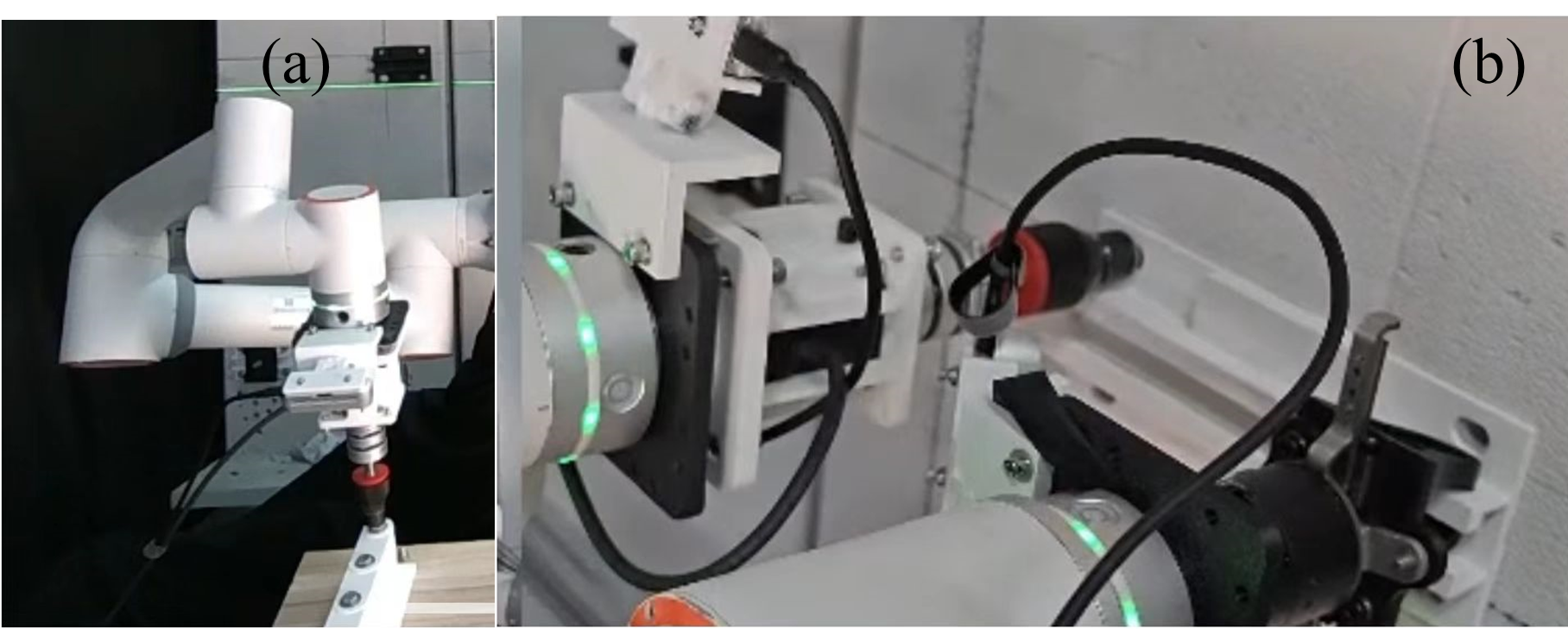}
    \caption{Dual-arm screw fastening task. The figure illustrates representative stages, including 
    (a) left-arm screw picking and 
    (b) screw fastening with the left arm while the right-arm gripper stabilizes the workpiece.}
    \label{fig:screw}
\end{figure}

\paragraph{Task 3: Dual-Arm Screw Fastening Task.}
As shown in Fig.~\ref{fig:screw}, the dual-arm screw fastening task requires coordinated manipulation of both robot arms. 
The task is decomposed into seven semantically distinct sub-tasks: 
\emph{(1) Left-Arm Screw Pick}, 
\emph{(2) Flip the Arm}, 
\emph{(3) Target Hole Identification}, 
\emph{(4) Coarse Approach}, 
\emph{(5) Fine Alignment}, 
\emph{(6) Screw Fastening}, and 
\emph{(7) Right-Arm Gripper Control}. 

In the semi-autonomous setting, the human operator assumes primary control of four sub-tasks—specifically, 
\emph{Left-Arm Screw Pick}, 
\emph{Coarse Approach}, 
\emph{Target Hole Identification}, and 
\emph{Right-Arm Gripper Control}—while the robot autonomously executes the remaining three sub-tasks. 
Applying the discrete SAR formulation, this allocation corresponds to an SAR of \(4/7 \approx 0.57\).

\rev{A trial is considered successful when the screw enters the target hole or
connection and remains engaged after screwdriver retraction without screw
dropping, jamming, safety-stop activation, or timeout. The achieved
insertion depth is evaluated separately as a continuous measure of fastening
quality.}

{\color{blue}
\begin{table}[htbp]
\centering
\caption{Performance of the three control modes in the nominal screw-fastening experiment. TeleOp and Semi-Auto each include 100 trials nested within 10 participants; Auto includes 100 unmanned pipeline repetitions. Continuous metrics are reported as trial-level mean $\pm$ sample standard deviation over successful trials; TeleOp metrics are based on 3 successful trials.}
\label{tab:screw_protocol}
\resizebox{\linewidth}{!}{%
\begin{tabular}{lccc}
\hline
Mode & Success (\%) & Insert. Depth (mm) & Time (s) \\
\hline
TeleOp    & 3 & $33.2 \pm 8.1$ & $231.4 \pm 45.2$ \\
Semi-Auto & 68 & $36.8 \pm 4.9$ & $109.6 \pm 16.8$ \\
Auto      & 93 & $48.9 \pm 2.7$ & $58.4 \pm 8.2$ \\
\hline
\end{tabular}
}
\end{table}
}

\rev{Pure teleoperation rarely succeeds due to the complexity of coordinated dual-arm manipulation. Semi-autonomous shared control shows higher observed success and shorter
execution time than TeleOp, while rule-based Auto achieves the best measured
performance under predefined conditions.}

\paragraph{Summary}
\rev{Across the three tasks, increasing robot responsibility improves task
completion, geometric accuracy, and execution efficiency under calibrated
laboratory conditions. Rule-based Auto achieves the highest measured
performance under these conditions, while Semi-Auto substantially improves
over TeleOp by assigning precision-critical subtasks to the robot and
retaining human guidance for target interpretation and coarse manipulation.
SAR describes the distribution of control responsibility, whereas subjective
operator workload is evaluated independently using Raw NASA-TLX.}

\rev{\subsection{Separation of Control Allocation and Subjective Workload}}
\rev{SAR and Raw NASA-TLX characterize different aspects of human involvement.
SAR describes how control responsibility is allocated between the operator
and the robot at the subtask level, whereas Raw NASA-TLX directly measures
the operator's perceived mental demand, physical demand, temporal demand,
performance, effort, and frustration. Accordingly, SAR is used only as a
control-allocation descriptor, while subjective workload is evaluated from
participant responses under TeleOp and Semi-Auto for the drilling task.}

\rev{\subsection{Drilling Robustness under Challenging Environmental Conditions}}
\label{sec:robustness_drilling}\label{sec:robustness_multi}

\rev{The nominal-condition results show that rule-based Auto performs best
when targets, lighting, and perception remain well calibrated. To evaluate
whether Semi-Auto provides greater adaptability when these assumptions are
disturbed, we conducted an additional robustness experiment using the
drilling task under challenging environmental conditions.

The nominal-reference rows in Table~\ref{tab:robust_drilling} reproduce the
nominal-condition drilling results previously reported in
Table~\ref{tab:drilling_protocol} and are included
only as reference baselines; they do not represent additional observations.
New trials were collected only for the disturbed environmental conditions.
The robustness comparison under disturbance focuses on rule-based Auto and
Semi-Auto. TeleOp is shown only in the nominal-reference rows because it was
not reevaluated under the disturbed conditions.}

\rev{\paragraph{Environmental conditions and protocol.}
Three categories of disturbance were evaluated. Lighting variation comprised
dim lighting and dynamic colored lighting. Partial occlusion covered 25\%,
50\%, and 75\% of the relevant target or visual region. Dynamic interference
comprised falling-object interference, robot-arm self-interference, and
interference from other moving objects in the workspace.

For each disturbed environmental condition, 100 new trials were performed
for Semi-Auto and 100 new trials were performed for rule-based Auto. The task
decomposition, human--robot responsibility allocation, skill sequence, and
success criteria were kept identical to those used in the corresponding
nominal-condition experiment.
Semi-Auto involved \(N{=}10\) operators, each completing 10 repetitions per
disturbed condition. Rule-based Auto remained unmanned and was evaluated
through 100 triggered-pipeline repetitions per disturbed condition. The
nominal-reference rows reproduce the previously reported nominal-condition results and
are provided only to facilitate comparison with the disturbed conditions.

Success rate is reported as the pooled proportion of trials satisfying the
binary task-completion criteria. Continuous outcomes are summarized as
trial-level mean \(\pm\) sample standard deviation over successful trials
within each control mode and disturbed environmental condition. Because the
continuous outcomes are conditional on task success, they are interpreted
together with the corresponding success rates.}

{\color{blue}
\begin{figure}[!htbp]
    \centering
    \includegraphics[width=\textwidth]{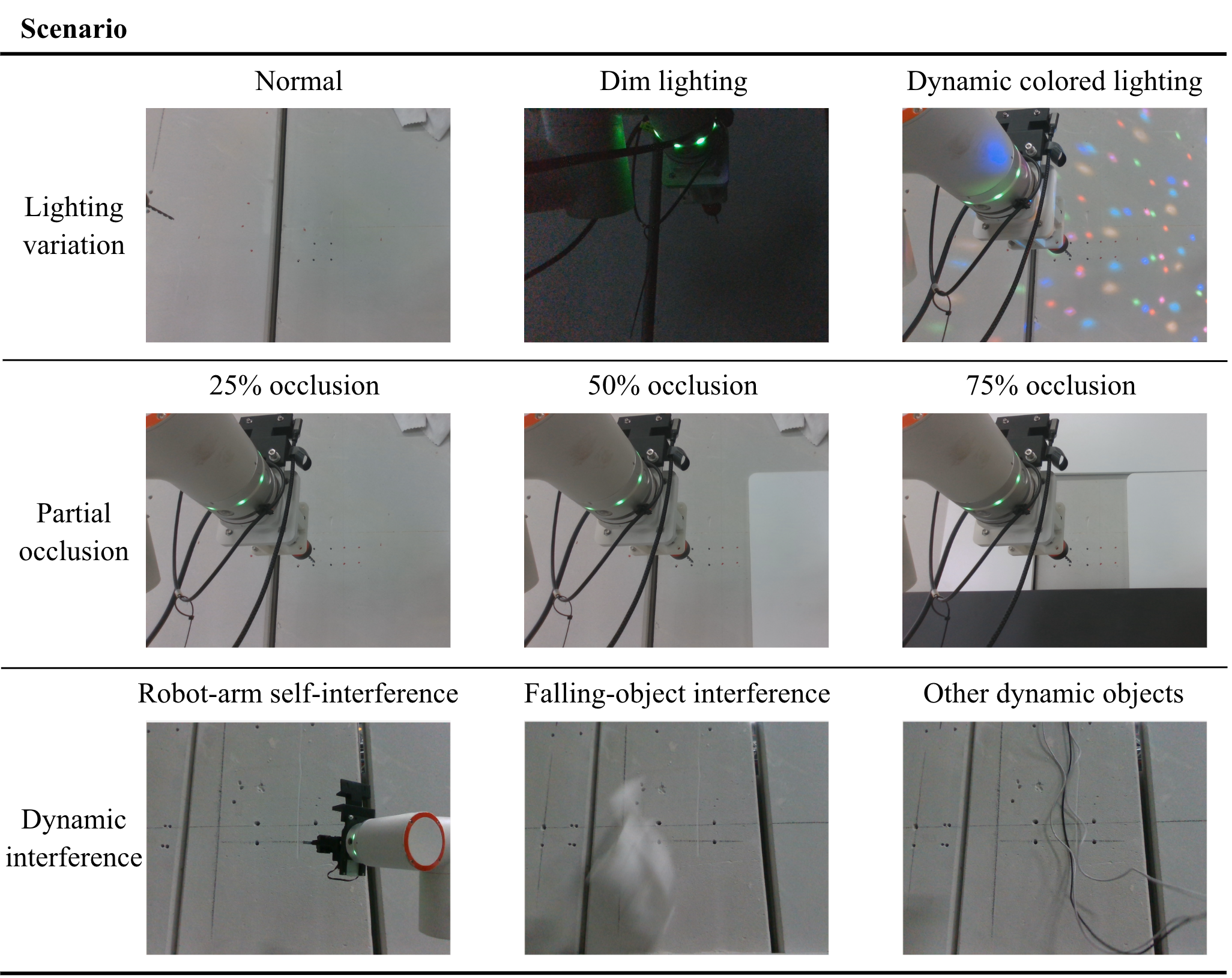}
    \caption{Representative environmental conditions used in the drilling robustness experiment. The evaluated conditions include normal lighting, dim lighting, dynamic colored lighting, 25\%, 50\%, and 75\% partial occlusion, robot-arm self-interference, falling-object interference, and interference from other dynamic objects.}
    \label{fig:challenging_scenarios}
\end{figure}
}

\rev{\paragraph{Compared modes and control allocation.}
The robustness experiment retained the same drilling-specific human--robot
responsibility allocation and execution procedure as the main Semi-Auto
and rule-based Auto conditions defined in
Sec.~\ref{sec:experimental_protocol} and
Table~\ref{tab:master_decomp}. No subtask was reassigned between the human
and the robot; only the environmental disturbance was varied.

Rule-based Auto executed the predefined perception and skill pipeline without
operator correction after the experimenter trigger. Semi-Auto retained the
same human--robot allocation used in the nominal experiment, while allowing
the operator to reinterpret the target, provide local pose corrections,
retry the currently active skill, or override the autonomous command. The
comparison therefore evaluates the robustness of the previously defined
control strategies under environmental disturbance rather than the effect of
changing the human--robot allocation.}

{\color{blue}
\begin{table}[!htbp]
\centering
\footnotesize
\caption{Performance of the drilling task under nominal and challenging environmental conditions. For each disturbed condition, Semi-Auto includes 10 operators with 10 trials each (100 trials), and Auto includes 100 unmanned repetitions. The nominal-reference rows reproduce the original nominal results. Continuous metrics are calculated over successful trials.}
\label{tab:robust_drilling}
\renewcommand{\arraystretch}{1.18}
\setlength{\tabcolsep}{4pt}
\resizebox{\textwidth}{!}{%
\begin{tabular}{@{}llccccc@{}}
\toprule
\textbf{Scenario} & \textbf{Mode} & \textbf{Success (\%)} & \textbf{Depth (mm)} & \textbf{Pos. Error (mm)} & \textbf{Perpendicularity ($^\circ$)} & \textbf{Time (s)} \\
\midrule
\multirow{3}{*}{\makecell[l]{Nominal\\reference}}
& TeleOp & 53 & $31.2 \pm 7.4$ & $36.4 \pm 9.2$ & $15.0 \pm 4.8$ & $92.5 \pm 18.6$ \\
& Semi-Auto & 82 & $50.6 \pm 2.3$ & $23.3 \pm 4.1$ & $2.2 \pm 0.7$ & $38.7 \pm 5.8$ \\
& Auto & 96 & $52.1 \pm 1.7$ & $15.0 \pm 2.5$ & $1.8 \pm 0.5$ & $15.0 \pm 2.0$ \\
\midrule
\multirow{2}{*}{Dim lighting}
& Semi-Auto & 80 & $50.4 \pm 2.4$ & $23.9 \pm 4.2$ & $2.3 \pm 0.8$ & $38.9 \pm 5.9$ \\
& Auto & 75 & $51.7 \pm 2.0$ & $25.9 \pm 5.2$ & $2.1 \pm 0.6$ & $15.6 \pm 2.2$ \\
\midrule
\multirow{2}{*}{\makecell[l]{Dynamic colored\\lighting}}
& Semi-Auto & 76 & $49.7 \pm 2.8$ & $25.4 \pm 4.9$ & $2.6 \pm 0.9$ & $40.2 \pm 6.4$ \\
& Auto & 63 & $50.7 \pm 2.5$ & $33.2 \pm 6.9$ & $2.4 \pm 0.8$ & $16.3 \pm 2.5$ \\
\midrule
\multirow{2}{*}{25\% occlusion}
& Semi-Auto & 81 & $50.5 \pm 2.3$ & $23.6 \pm 4.0$ & $2.3 \pm 0.7$ & $38.8 \pm 5.8$ \\
& Auto & 91 & $51.8 \pm 1.8$ & $15.8 \pm 2.7$ & $2.2 \pm 0.6$ & $15.5 \pm 2.1$ \\
\midrule
\multirow{2}{*}{50\% occlusion}
& Semi-Auto & 77 & $49.8 \pm 2.8$ & $24.8 \pm 4.8$ & $3.0 \pm 1.0$ & $39.8 \pm 6.3$ \\
& Auto & 56 & $50.8 \pm 2.6$ & $17.1 \pm 3.2$ & $6.5 \pm 2.1$ & $16.0 \pm 2.4$ \\
\midrule
\multirow{2}{*}{75\% occlusion}
& Semi-Auto & 73 & $49.1 \pm 3.1$ & $25.5 \pm 5.1$ & $3.4 \pm 1.2$ & $40.8 \pm 6.8$ \\
& Auto & 33 & $49.3 \pm 3.2$ & $18.0 \pm 3.6$ & $11.2 \pm 3.5$ & $16.8 \pm 2.8$ \\
\midrule
\multirow{2}{*}{\makecell[l]{Falling-object\\interference}}
& Semi-Auto & 82 & $50.6 \pm 2.3$ & $23.5 \pm 4.1$ & $2.2 \pm 0.7$ & $38.9 \pm 5.8$ \\
& Auto & 93 & $51.8 \pm 1.9$ & $15.7 \pm 2.6$ & $2.1 \pm 0.6$ & $15.4 \pm 2.1$ \\
\midrule
\multirow{2}{*}{\makecell[l]{Robot-arm\\self-interference}}
& Semi-Auto & 75 & $49.1 \pm 3.0$ & $26.0 \pm 5.0$ & $3.2 \pm 1.1$ & $41.0 \pm 6.7$ \\
& Auto & 47 & $47.1 \pm 3.9$ & $30.9 \pm 7.0$ & $7.6 \pm 2.7$ & $18.0 \pm 3.4$ \\
\midrule
\multirow{2}{*}{\makecell[l]{Other dynamic\\objects}}
& Semi-Auto & 74 & $48.8 \pm 3.2$ & $26.8 \pm 5.3$ & $3.4 \pm 1.2$ & $41.3 \pm 6.9$ \\
& Auto & 42 & $46.2 \pm 4.3$ & $33.8 \pm 7.6$ & $8.5 \pm 3.0$ & $18.7 \pm 3.7$ \\
\bottomrule
\end{tabular}%
}
\end{table}
}

\rev{\paragraph{Results and interpretation.}
The results reveal a condition-dependent trade-off. Under nominal conditions and mild transient disturbances, including 25\% occlusion and falling-object interference, rule-based Auto retains the highest success, accuracy, and efficiency. These cases do not substantially disrupt the predefined perception and execution pipeline. Under stronger or persistent disturbances, including dim or dynamic colored lighting, 50--75\% occlusion, robot-arm self-interference, and other moving objects, Auto success decreases sharply and drilling error increases. Semi-Auto remains closer to its nominal success level because the operator can reinterpret the scene, correct the target or local pose, and re-trigger autonomous skills.}

\rev{Completion-time and continuous-accuracy values in Table~\ref{tab:robust_drilling} are conditional on successful trials. Consequently, the shorter Auto times under severe disturbances do not account for failed trials and should not be interpreted as superior overall robustness. The success-rate results provide the primary evidence for the robustness comparison. The findings therefore do not show that Semi-Auto universally outperforms Auto; rather, Auto is preferable for calibrated and mildly disturbed drilling, whereas Semi-Auto provides greater adaptability when persistent environmental interference exceeds the predefined pipeline's correction capability.}

\rev{\subsection{Subjective Workload Evaluation}}
\rev{Subjective workload was evaluated using Raw NASA-TLX for TeleOp and
Semi-Auto in the drilling task. Ten participants evaluated each
mode in terms of mental demand, physical demand, temporal
demand, perceived performance, effort, and frustration
~\cite{Hart2006_NASATLX}. The performance scale was coded so that a higher
value indicates poorer perceived performance; thus, lower scores indicate
lower workload for all six dimensions. For each questionnaire record and drilling-mode condition,
the Raw TLX score was calculated as the unweighted mean of the six
dimensions. Table~\ref{tab:nasa_tlx_by_task} reports the mean and sample
standard deviation across the ten participant-level scores.}

{\color{blue}
\begin{table}[!htbp]
\centering
\footnotesize
\caption{Subjective workload under TeleOp and Semi-Auto control for the drilling task, measured using the Raw NASA-TLX. Values are mean $\pm$ sample standard deviation across ten participants. Lower scores indicate lower subjective workload.}
\label{tab:nasa_tlx_by_task}
\renewcommand{\arraystretch}{1.2}
\setlength{\tabcolsep}{4pt}
\resizebox{\textwidth}{!}{%
\begin{tabular}{llccccccc}
\toprule
\textbf{Task} & \textbf{Mode} & \textbf{Mental} & \textbf{Physical} & \textbf{Temporal} & \textbf{Performance} & \textbf{Effort} & \textbf{Frustration} & \textbf{Raw TLX} \\
\midrule
\multirow{2}{*}{Drilling}
& TeleOp & $67.5 \pm 9.5$ & $52.5 \pm 9.2$ & $62.5 \pm 8.2$ & $57.5 \pm 10.1$ & $70.0 \pm 9.1$ & $60.0 \pm 8.8$ & $\mathbf{61.7 \pm 8.0}$ \\
& Semi-Auto & $40.0 \pm 7.5$ & $27.5 \pm 8.6$ & $35.0 \pm 7.8$ & $27.5 \pm 9.5$ & $37.5 \pm 7.2$ & $30.0 \pm 8.5$ & $\mathbf{32.9 \pm 6.8}$ \\
\bottomrule
\end{tabular}%
}
\end{table}
}

\rev{The observed mean Raw TLX score for drilling is lower under Semi-Auto than under TeleOp, decreasing from 61.7 to 32.9, a relative reduction of 46.7\%. Lower observed mean scores under Semi-Auto are found across all six workload dimensions. These results indicate a consistent descriptive workload trend in favor of Semi-Auto for the evaluated drilling task.}

\subsection{\rev{Drilling Task: Subtask Allocation Ratio Analysis}}

\rev{This subsection reports a dedicated drilling-only SAR allocation sweep with fixed subtask assignments (Semi-Auto 1--4). It is separate from the main three-mode comparison in Table~\ref{tab:drilling_protocol}. Although the main Semi-Auto protocol also uses a 3/5 human allocation (SAR $=0.6$), the two tables correspond to different experimental protocols: the main comparison allows flexible operator recovery within the shared-control policy, whereas the SAR sweep enforces fixed stage-wise allocations for controlled analysis. Human-operated SAR levels (SAR $>0$) were collected with the same \(N{=}10\) operators (10 trials per operator per SAR level); the Auto row (SAR $=0$) reports \(N{=}100\) unmanned pipeline repeats matched to the main Auto baseline and does not require an operator during execution.}

\rev{The drilling SAR sweep evaluates a nested sequence of fixed stage-wise
role allocations across Target Selection, Coarse Approach, Fine Alignment,
Drilling, and Retraction. Because both the SAR value and the identity of the
human-controlled subtasks change across conditions, the sweep is interpreted
as a descriptive sensitivity analysis of role allocation rather than a
full-factorial component ablation. The independent contributions of the VR
interface, autonomous fine alignment, the Flip-the-Arm procedure, and
façade-plane estimation were not isolated and remain for future
investigation.}

\begin{figure}[htbp]
    \centering
    \includegraphics[width=1\linewidth]{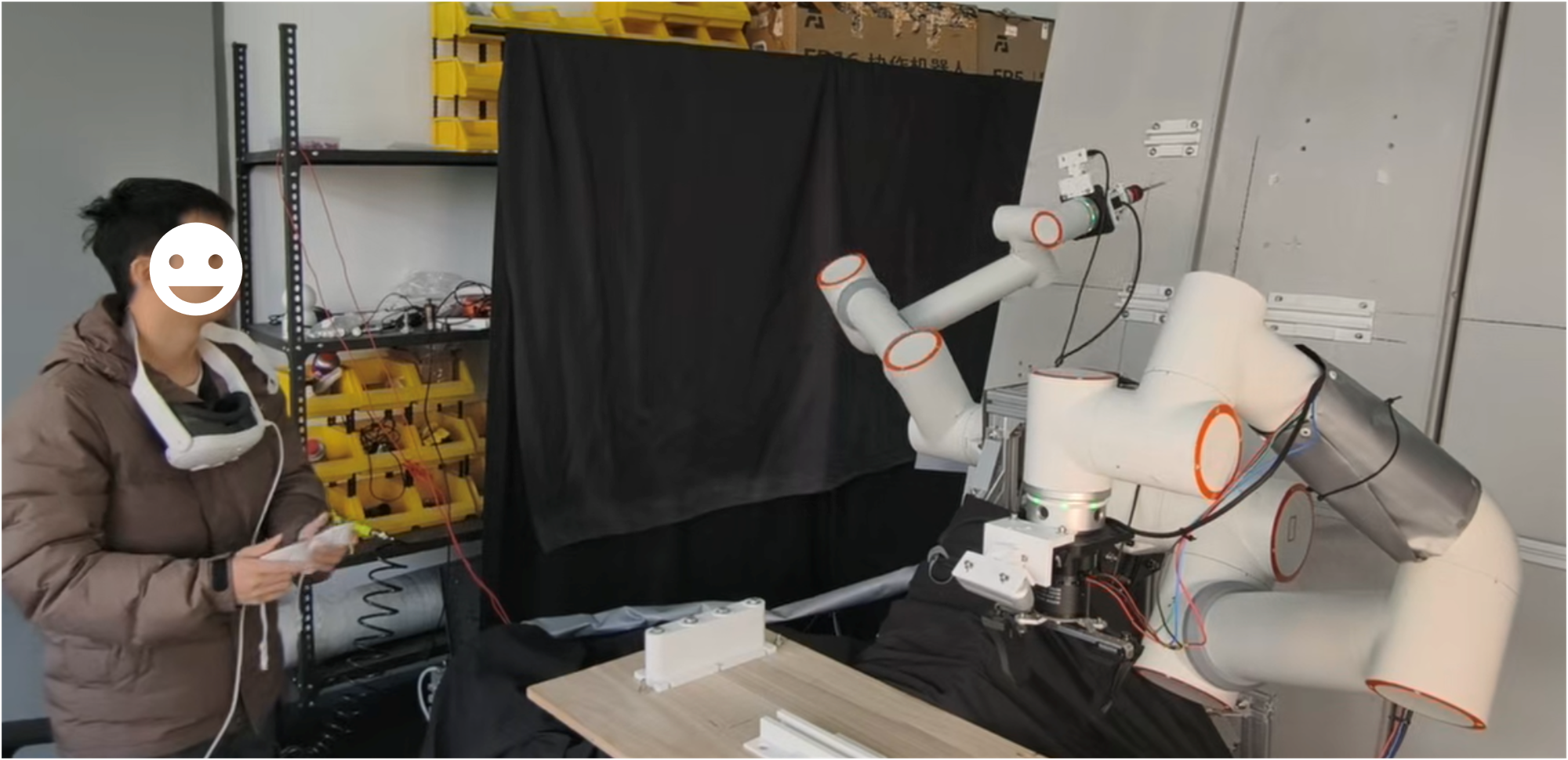}
    \caption{Teleoperated drilling process. A human operator used VR devices to control the robotic arm during drilling.}
    \label{fig:SARdrill}
\end{figure}

As shown in Fig.~\ref{fig:SARdrill}, the drilling task is executed under human teleoperation during the drilling process. 
For analysis, the task is decomposed into five sequential sub-tasks, corresponding to the following skill primitives:
\emph{Target Selection} ($\mathcal{S}_1$),
\emph{Coarse Approach} ($\mathcal{S}_2$),
\emph{Fine Alignment} ($\mathcal{S}_3$),
\emph{Drilling} ($\mathcal{S}_4$), and
\emph{Retraction} ($\mathcal{S}_5$).
\rev{A trial is considered successful when the drill penetrates the wall and
produces a measurable hole without jamming, safety-stop activation, or
timeout. Drilling depth, positional error, and perpendicularity are evaluated
separately as continuous performance measures.}

For each skill primitive, the execution strategy varies depending on whether control is human-dominated or autonomous.
During \emph{Target Selection} ($\mathcal{S}_1$), human-dominated control requires the operator to visually identify the target location, whereas autonomous execution uses a depth-camera-based Hole Detection algorithm to determine the target hole position $p_h$ on the facade, ensuring stable and accurate localization.
During \emph{Coarse Approach} ($\mathcal{S}_2$), human-dominated control depends on the operator manually guiding the end-effector toward the target region, while autonomous execution follows a planned trajectory derived from the predefined target pose.
During \emph{Fine Alignment} ($\mathcal{S}_3$), human-dominated control relies on the operator’s visual judgment to correct minor pose errors, whereas autonomous control performs precise alignment using depth-camera-based Facade Plane Estimation and Pose Alignment.
During \emph{Drilling} ($\mathcal{S}_4$), human-dominated control requires continuous regulation of tool motion and contact force by the operator, while autonomous execution follows a predefined drilling policy with fixed motion parameters.
During \emph{Retraction} ($\mathcal{S}_5$), human-dominated control involves manually withdrawing the tool from the drilled hole, whereas autonomous control executes a predefined retraction trajectory to safely disengage from the contact surface.

Human involvement in each skill is represented using a binary control label
\(\delta_i \in \{0,1\}\), where \(\delta_i = 1\) indicates that the skill is primarily human-controlled, and \(\delta_i = 0\) indicates that the skill is executed autonomously by the robot. 
The discrete Subtask Allocation Ratio (SAR) is defined in Eq.\eqref{SAR}
where \(N=5\) for the drilling task. SAR values closer to 1 indicate predominantly human-driven task execution, whereas values closer to 0 correspond to largely autonomous execution. The discrete SAR can be interpreted as a special case of the continuous human-control weight \(\alpha_i(t)\), where \(\alpha_i(t)\) is binarized for each skill primitive.

Table~\ref{tab:sar_modes} summarizes the binary allocation of human control across the five skills for each control mode. The table header explicitly maps each skill index to its semantic meaning.

\begin{table}[htbp]
\centering
\caption{\rev{Discrete subtask allocation under different control modes for the dedicated drilling SAR sweep ($N{=}5$ skills).}}
\label{tab:sar_modes}
\begin{tabular}{l p{7.2cm} c}
\hline
Control Mode 
& Binary Control Vector $\boldsymbol{\delta} = (\delta_1,\dots,\delta_5)$ for Task $\mathcal{S}_i$
& SAR \\
\hline
Auto        
& $(0,0,0,0,0)$ 
& 0.0 \\

Semi-Auto 1 
& $(1,0,0,0,0)$ 
& 0.2 \\

Semi-Auto 2 
& $(1,1,0,0,0)$ 
& 0.4 \\

Semi-Auto 3 
& $(1,1,1,0,0)$ 
& 0.6 \\

Semi-Auto 4 
& $(1,1,1,1,0)$ 
& 0.8 \\

TeleOp      
& $(1,1,1,1,1)$ 
& 1.0 \\
\hline
\end{tabular}
\end{table}

The corresponding mean performance metrics for each control mode are shown in Table~\ref{tab:drilling_comprehensive_sar}. All values are presented as averages.

\begin{table}[htbp]
\centering
\footnotesize
\caption{\rev{Mean performance in the dedicated drilling SAR sweep (separate from Table~\ref{tab:drilling_protocol}). Human-operated rows: \(N{=}100\) participant trials each; Auto: \(N{=}100\) unmanned repeats.}}
\label{tab:drilling_comprehensive_sar}
\resizebox{\textwidth}{!}{%
\begin{tabular}{lcccccc}
\toprule
Control Mode & SAR & Success (\%) & Depth (mm) & Positional Error (mm) & Perpendicularity ($^\circ$) & Time (s) \\
\midrule
Auto        & 0.0 & 96 & 52.1 & 15.0 & 1.8  & 15.0 \\
Semi-Auto 1 & 0.2 & 92 & 51.3 & 16.8 & 2.5  & 18.3 \\
Semi-Auto 2 & 0.4 & 88 & 50.9 & 17.2 & 3.0  & 21.7 \\
Semi-Auto 3 & 0.6 & 72 & 50.6 & 20.3 & 12.2 & 32.4 \\
Semi-Auto 4 & 0.8 & 58 & 49.8 & 26.5 & 14.5 & 58.5 \\
TeleOp      & 1.0 & 53 & 31.2 & 36.4 & 15.0 & 92.5 \\
\bottomrule
\end{tabular}%
}
\end{table}

\paragraph{Results and Discussion.}
\rev{The performance metrics in Table~\ref{tab:drilling_comprehensive_sar} demonstrate a monotonic relationship between SAR and task outcomes in the controlled drilling sweep. Rule-based Auto (SAR $=0$) achieves the highest success rate, consistent drilling depth, minimal angular deviation, and the shortest execution time under predefined conditions.}

As SAR increases from 0.2 to 0.8, performance progressively degrades across all metrics. Moderate human involvement (SAR $\leq 0.4$) preserves relatively high success rates and acceptable accuracy, whereas assigning the human operator to precision-critical skills such as fine alignment and drilling (SAR $\geq 0.6$) results in substantial increases in positional error, perpendicularity deviation, and execution time. This trend highlights the difficulty for novice operators to manually perform contact-rich, high-precision 6-DoF manipulations.

Full teleoperation (SAR $=1$) exhibits the lowest overall performance, with reduced success rate, shallow and variable drilling depth, poor perpendicularity, and substantially prolonged task duration. These results indicate that fully manual execution is ill-suited for precision-intensive manipulation tasks, especially for inexperienced operators.

\rev{Overall, discrete SAR provides an interpretable framework for analyzing human--robot role allocation. Limiting human involvement to coarse subtasks can improve outcomes relative to full teleoperation under nominal conditions, but rule-based Auto remains numerically best when calibration holds. Drilling robustness under challenging environmental conditions is evaluated separately in Sec.~\ref{sec:robustness_drilling}.}

\rev{\subsection{Failure Mode Analysis}}
\rev{TeleOp failures were primarily associated with operator-alignment errors and kinematic or reachability difficulties, particularly during complex manual positioning and arm-reorientation stages. Semi-Auto failures mainly arose from perception and localization errors, residual operator-alignment errors, and contact jamming. Auto failures were chiefly related to perception or localization errors, target mismatch, contact jamming, and kinematic limitations. These observations distinguish the characteristic failure mechanisms of the three control modes rather than relying only on aggregate success rates.}

\rev{\subsection{Limitations and Future Work}}
\rev{The proposed framework was evaluated in a laboratory façade mock-up rather than in a high-altitude field environment. The nominal experiments covered drilling, bracket pick-and-place, and screw fastening, whereas the environmental-robustness evaluation was limited to drilling under lighting variation, 25--75\% partial occlusion, falling-object interference, robot-arm self-interference, and interference from other moving objects. The robustness evaluation did not include target displacement, façade-plane tilt, controlled depth-noise injection, or robustness testing of bracket placement and screw fastening.

The NASA-TLX evaluation was limited to the drilling task and did not assess
subjective workload for bracket placement or screw fastening. The human-operated outcomes are interpreted descriptively as system-level
validation within the evaluated participant group and experimental
conditions; population-level inference regarding operator effects is beyond
the scope of the present study. Future larger-cohort evaluations will include
participant-level aggregation and repeated-measures modeling.
Objective workload indicators, including override frequency, VR-controller
path length, physiological responses, and long-duration fatigue, were not
quantified.

Additional limitations include: (i) the absence of full-factorial ablations isolating the contributions of the VR interface, autonomous fine alignment, the automated arm-reorientation
procedure, and façade-plane estimation; (ii) the absence of comparisons with joystick- or 3D-mouse-based teleoperation and learning-from-demonstration baselines; (iii) the use of a fixed rule-based skill library without progressive online skill discovery; and (iv) the absence of direct high-altitude field validation. Future work will broaden the evaluated disturbance families, incorporate objective and longitudinal human-factor measures, and conduct component-level ablations. Learning-based extensions will build on the present skill framework by using VR teleoperation trajectories as demonstrations for imitation learning, shared-control interventions as corrective labels, reinforcement learning to optimize contact-rich skill primitives under explicit safety constraints, and multimodal perception to improve target localization under occlusion and lighting variation. These extensions will retain operator override and safety supervision while reducing dependence on predefined targets and fixed execution pipelines.}


\section{Conclusion}

\rev{This work introduces a human-in-the-loop control paradigm for construction robotics that enables structured transitions between VR-guided teleoperation, semi-autonomous shared control, and rule-based autonomous dual-arm execution under predefined laboratory conditions. The dual-arm system integrates immersive VR guidance with a manually defined rule-based skill library for multi-stage, contact-rich façade tasks, while a shared-autonomy framework supports switching between human input and autonomous execution with explicit command-priority rules. Under nominal conditions and mild disturbances, rule-based Auto achieves the highest measured success, accuracy, and efficiency when targets and perception are well calibrated. The drilling robustness experiment shows that under severe lighting degradation, 50--75\% occlusion, robot-arm self-interference, and other persistent dynamic interference, Semi-Auto maintains higher success by allowing the operator to reinterpret the scene and provide corrective guidance. The drilling-task Raw NASA-TLX results further show that the observed mean subjective workload under Semi-Auto is 46.7\% lower than under TeleOp. Semi-Auto is therefore advantageous for workload reduction in the evaluated drilling task and for uncertainty recovery rather than nominal speed. These results support conditional control-strategy selection: rule-based Auto for well-calibrated repetitive operations, and Semi-Auto for façade tasks where environmental uncertainty or sustained human involvement may require shared control.}

\bibliographystyle{elsarticle-num} 
\bibliography{reference}

\end{document}